# A Large Language Model-based multi-agent manufacturing system for intelligent shopfloors

*Zhen Zhao[a], Dunbing Tang[a,*], Changchun Liu[a,b,*], Liping Wang[a], Zequn Zhang[a], Haihua Zhu[a], Kai Chen[a], Qingwei Nie[c], Yuchen Ji[a],*

*[a]College of Mechanical and Electrical Engineering, Nanjing University of Aeronautics and Astronautics, Nanjing 210016, People's Republic of China*

*[b]Key Laboratory of System Control and Information Processing, Ministry of Education of China, Shanghai 200240, People's Republic of China*

*[c]College of Mechanical Engineering, Yangzhou University, Yangzhou 225127, People's Republic of China*

**Abstract:** As customer demand for multi-variety and small-batch production increases, dynamic disturbances place greater demands on manufacturing systems. To address such challenges, researchers proposed the multi-agent manufacturing system. However, conventional agent negotiation typically relies on pre-defined and fixed heuristic rules, which are ill-suited to managing complex and fluctuating disturbances. In current implementations, mainstream approaches based on reinforcement learning require the development of simulators and training models specific to a given shopfloor, necessitating substantial computational resources and lacking scalability. To overcome this limitation, the present study proposes a Large Language Model-based (LLM-based) multi-agent manufacturing system for intelligent shopfloor management. By defining the diverse modules of agents and their collaborative methods, this system facilitates the processing of all workpieces with minimal human intervention. The agents in this system consist of the Machine Server Module (MSM), Bid Inviter Module (BIM), Bidder Module (BM), Thinking Module (TM), and Decision Module (DM). By harnessing the reasoning capabilities of LLMs, these modules enable agents to dynamically analyze shopfloor information and select appropriate processing machines. The LLM-based modules, predefined by system prompts, provide dynamic functionality for the system without the need for pre-training. Extensive experiments were conducted in physical shopfloor settings. The results demonstrate that the proposed system exhibits strong adaptability, and achieves superior performance (makespan) and stability (as measured by sample standard deviation) compared to other approaches without requiring pre-training.

## 1 Introduction

Mass personalization is becoming increasingly prevalent as productivity improves[1]. With this shift, customer demand for unique, customized products is growing more frequent. The need for multi-variety, small-batch production drives constant changes in manufacturing resources, placing greater emphasis on the manufacturing system's ability to efficiently organize and manage these resources in response to evolving demand.

Manufacturing systems serve the purpose of organizing manufacturing resources on shopfloors or larger areas for efficient production. Traditional production approaches require production schedulers to coordinate workpieces based on their expertise and real-time conditions on the shopfloor. Manual scheduling often involves collaboration across multiple machines and departments, with schedules typically set over extended periods. These schedules are generally rigid and resistant to change. This rigidity is ill-suited to accommodate the fluctuating demands of modern production. Conventional manufacturing systems are optimized for large-scale production of standardized, uniform workpieces. However, the growing need for personalized, custom products requires multi-variety and small-batch production, for which building dedicated production lines is both uneconomical and inefficient. In this context, flexible manufacturing offers a more viable solution. Specifically, the framework to solve Flexible Job-shop Scheduling Problem (FJSP) provides a way for organizing product processing through adaptive scheduling methods, enabling the efficient production of complex and variable products.

The conventional scheduling methods, typically represented by metaheuristic algorithms[2,3], can effectively solve the static FJSP problem. However, these methods require continuous re-scheduling to adapt to changing scenarios, making them computationally intensive due to the iterative calculations involved. To address this problem, the multi-agent manufacturing system is proposed. In this system, manufacturing resources are treated as distinct agents, which are coordinated and managed through negotiation processes among them. In contrast to metaheuristic algorithms, the processing machine for each workpiece in multi-agent manufacturing system using reflection-scheduling to address dynamic FJSP, which determined the processing machine only after the previous processing step is completed. This approach, where machine assignment is not pre-determined, offers flexibility by allowing decisions to be made based on the real-time conditions of the shopfloor. Under the architecture of a multi-agent manufacturing system, the question of optimization shifts towards discovering an intelligent negotiation mechanism.

Conventional negotiation mechanisms in multi-agent manufacturing systems mainly rely on heuristic rules. These rules, designed by human expertise, provide rapid responses but lack the intelligence required to select the optimal machine. Deep Reinforcement Learning (DRL) algorithms are introduced to address this problem [4–6]. DRL algorithms provide a more efficient method, capable of swiftly seeking scheduling solutions and dealing with dynamic disturbances. These methods can effectively integrate diverse information about orders and machines. Nonetheless, with the combination of multi-agent reinforcement learning and multi-agent manufacturing system, the Partially Observable Markov Decision Process (POMDP) is also introduced, which poses a challenge to the stability of these algorithms. Moreover, even though the transfer learning is introduced by some researches, pre-training is still required by DRL, implying that such algorithms cannot be deployed on a new shopfloor swiftly. Simultaneously, while DRL demonstrate strong performance in small-scale manufacturing resource scheduling, their effectiveness diminishes as the scale of manufacturing resources increases. This decline in performance is attributed to the rapid increase in algorithmic complexity. As such, the key challenges lie in improving the scalability and minimizing solution time when seeking optimal scheduling solutions, all while maintaining a high level of precision.

Large Language Models (LLMs) offer a promising way to enhance the capabilities of agents in the multi-agent manufacturing system. LLMs, exemplified by ChatGPT, have sparked a fresh wave of revolution in Artificial Intelligence (AI) [7]. Through its training process, LLMs acquire a vast amount of textual data, endowing them with strong, human-like language generation capabilities. It can be anticipated that replacing heuristic rules with LLMs to enhance the intelligence of agents offers a promising solution. To address the challenges of improving the efficiency and reducing the complexity of manufacturing resource scheduling, this study proposes an LLM-based multi-agent manufacturing system. Unlike metaheuristic algorithms, which rely on time-consuming iterative calculations, and DRL algorithms, which require pre-training, the proposed system eliminates these limitations. Instead, it enables dynamic goal-setting and adjustments through the design of prompts, akin to conversing with a human. Further, this system can be rapidly deployed on new physical shopfloors. The main contributions of the present study can be summarized as follows:

Table 1 The acronyms and their definition in this paper

| **Acronyms** | **Definition** |
|---|---|
| LLM | Large Language Model |
| DRL | Deep Reinforcement Learning |
| MSM | Machine Server Module |
| BIM | Bid Inviter Module |
| BM | Bidder Module |
| TM | Thinking Module |
| DM | Decision Module |
| FJSP | Flexible Job-shop Scheduling Problem |
| AI | Artificial Intelligence |
| IIOT | Industrial Internet of Things |
| IPC | Industrial Personal Computer |
| PLC | Programmable Logic Controller |
| API | Application Programming Interface |
| Markdown format | A markup language for creating formatted text using a plain-text editor |
| Makespan | The length of time that elapses from the start of work to the end |
| CoT | Chain of Thought, a technique that allows LLMs to solve a problem as a series of intermediate steps before giving a final answer |

(1) The system assigns diverse modules to each agent of manufacturing resource and defines their LLM-based collaborative methods. With the support of LLM-based modules, the negotiation among agents avoids the drawback that a single heuristic rule cannot choose a suitable machine promptly according to the current shopfloor situation. Agents can negotiate the overall processing task based on the production task using natural language, which is different from other scheduling methods. Shopfloor leaders can integrate and utilize LLMs through a straightforward dialogue, thereby customizing the system to align with their individual objectives.

(2) In the proposed system, both the data collection and training processes typical of conventional AI methods are avoided, significantly reducing the complexity of scheduling. Because of the flexibility and autonomy of the LLM-based agents, the system can be quickly adapted to the target manufacturing scenario without requiring specific reconfiguration, while still delivering better performance than conventional methods.

(3) Far from being confined to theoretical exploration, the agent in this system utilizes an MSM to operate the manufacturing resources. With the support of MSM, the LLM-based agent can directly regulate the orders of manufacturing resources, even autonomously executing the complete processing cycle of a product by negotiation between agents without human interference.

The remainder of this paper is organized as follows. Section 2 reviews related works in the field. Section 3 presents a detailed discussion of the LLM-based multi-agent manufacturing system. Section 4 provides an in-depth description of the agents and their modules introduced in Section 3. Section 5 outlines the experimental setup and compares the performance of the proposed system with that of traditional heuristic rule-based approaches. Finally, Section 6 summarizes the key findings and contributions of this study. In order to increase the readability, the acronyms utilized in the present study are summarized and presented in Table 1.

## 2 Related work

The analysis and inference capabilities of LLMs are set to introduce new levels of intelligence into manufacturing systems. As the Industrial Internet of Things (IIoT) becomes increasingly prevalent, there is a growing demand for enhanced intelligence within manufacturing environments. To meet this need, current scheduling approaches predominantly rely on metaheuristic and DRL algorithms. This section provides an overview of existing efforts related to scheduling methods in manufacturing systems, as well as some applications of LLMs in this context.

### 2.1 Scheduling methods of the manufacturing system

For an extended period, researchers have concentrated on the scheduling problem in the manufacturing system. Graey et al. [8] demonstrated that the shortest-length schedule and minimum mean-flow-time schedule in flow-shop scheduling is NP-complete. This insight steered research away from seeking optimal solutions and toward finding acceptable, practical solutions for flow-shop scheduling. Similar to the traveling salesman problem, the focus of scheduling research has shifted from obtaining exact mathematical solutions to identifying feasible and efficient ones.

The metaheuristic algorithm is introduced to address this problem. Jian et al. [9] proposed a cloud edge-based two-level hybrid scheduling learning model and improved long and short-term memory networks model is put forward for fast prediction. Liu et al. [10] formulated a mathematical model that aims to optimize the minimum production cycle for the dual-resource batch scheduling in a flexible job shop. To address this issue, they developed an enhanced nested optimization algorithm, whose efficacy has been substantiated through the examination of real-world scenarios. For the purpose of addressing the scheduling challenge within a flexible job shop that utilizes segmented automatic guided vehicles, Liu et al. [11] developed a dual-resource optimization model for machine tools and automatic guided vehicles, with the objective of minimizing the makespan. This study introduced an enhanced genetic algorithm tailored to resolve the aforementioned problem. Concurrently, Li et al. [12] introduced an innovative, adaptive memetic algorithm that draws upon popularity-based principles. This algorithm was designed to rectify certain shortcomings and is applied to the energy-efficient distributed flexible job shop scheduling problem, with the dual objectives of minimizing both the makespan and energy consumption. The scheduling methods based on the metaheuristic algorithm demonstrate high precision but require additional time to compute the solution. Consequently, this kind of method is effectively suitable for static scheduling problems within manufacturing systems.

To solve the dynamic scheduling problems, researchers have conducted extensive studies based on DRL. Liu et al. [13] proposed a predictive maintenance approach for machine tools for DRL approaches to extract features in shopfloor. Gui et al. [5] proposed a DRL approach to minimize the mean tardiness, which selected the most appropriate weights for dispatching rules. An AI-based scheduling system that employs composite reward functions was introduced by Zhou et al. [14]. This system was designed for data-driven dynamic scheduling of manufacturing jobs within the context of a smart factory, where uncertainty is a factor. This scheduler demonstrated the ability to enhance multi-objective performance metrics associated with production scheduling challenges. Du et al. [15] proposed a DQN model to solve a multi-objective flexible job shop problem with crane transportation and setup times. Considering the complexity of this problem, this study also designed an identification rule to organize the crane transportation in solution decoding. Liu et al. [16] proposed a hierarchical and distributed architecture to solve the dynamic flexible job shop scheduling problem to facilitate real-time control. Luo [17] proposed a deep Q-network to cope with continuous production states and learn the most suitable action at each rescheduling point. Wang et al. [18] proposed a scheduling algorithm that is tailored to address job scheduling problems within a resource preemption context, leveraging multi-agent reinforcement learning. Chen et al. [19] introduced a self-learning genetic algorithm framework, which utilizes the genetic algorithm as its foundational optimization technique, with its pivotal parameters being intelligently tuned through DRL. This work merges these two algorithms utilizing DRL in conjunction with the meta-heuristic method to address dynamic disturbance issues.

Nevertheless, the effectiveness of DRL-based algorithms in solving dynamic scheduling problems arises from their training simulator, which also limits their performance and scalability. The multi-agent manufacturing system is increasingly emerging as a prominent solution. Qin et al. [20] conducted a comprehensive review of the literature on self-organizing manufacturing systems and introduced a comprehensive concept of self-organizing manufacturing networks. This concept is positioned as the next evolutionary step in manufacturing automation technologies, specifically aimed at facilitating mass personalization. Building upon this, Qin et al. [1] developed a reinforcement learning-based approach that combines static training with dynamic execution. This approach is designed to address dynamic job shop scheduling issues within the framework of a self-multi-agent manufacturing network. Additionally, Alexopoulos et al. [21] designed a framework for the modeling and deployment of a DRL agent to support short-term production scheduling. With the minimizing the production makespan, their DRL agent can learn the suitable dispatching policy. Kim et al. [6] introduced a smart manufacturing system that employs a multi-agent system and reinforcement learning. This system is distinguished by its use of intelligent agents embedded within machines, which enable the system with autonomous decision-making capabilities, the ability to interact with other systems, and the intelligence to adapt to dynamically changing environments. Wang et al. [22] proposed a smart factory framework that integrates industrial networks, cloud technology, and supervisory control terminals with smart shop-floor objects. This framework leverages the feedback and coordination by the central coordinator in order to achieve high efficiency. Gui et al. [4] introduced the DRL into multi-agent manufacturing system, to solves the dynamic FJSP with the objective of minimizing the mean tardiness. Based on this, their work achieved excellent performance while maintaining scalability. Qin and Lu [23] proposed knowledge graph-enhanced DRL method within that combines domain knowledge from historical production records with adaptive scheduling policies. Their approach has shown faster learning rates compared to traditional DRL, while still needing training.

The multi-agent manufacturing system is characterized by its swift processing speed and obviates the need for pre-training, thereby serving as an effective procedure for migrating and augmenting the manufacturing system. Nevertheless, the traditional negotiations among agents within this system cannot change their policy of machine selection based on the real-time conditions of the shopfloor, which still requires completion. Neither single DRL methods or DRL-based metaheuristic methods can make decisions based on real-time conditions without a pre-training phase. Ensuring dynamic decision-making capabilities for agents within multi-agent manufacturing systems, while preserving scalability, has emerged as a critical challenge.

### 2.2 Applications of LLMs

Transformer [24] has emerged as a groundbreaking, versatile technique in Natural Language Processing (NLP), particularly within the context of LLMs. As computing power and data availability have grown, so too have the capabilities of LLMs, such as the GPT series [25–28]. Notably, the GPT-3.5 version, popularly known as ChatGPT [29], marked a significant milestone with its introduction of multimodal functionality and highly realistic conversational abilities.

LLMs have demonstrated their capabilities in various fields. In biology, researchers have achieved immense progress building upon the use of LLMs. Boiko et al. [30] introduced an AI system powered by GPT-4. This system is capable of autonomously designing, planning, and executing intricate experiments. In the field of chemistry, Jablonka et al. [31] fine-tuned GPT-3 to answer chemical questions in natural language with the correct answer. Researchers have also invested significant efforts into enhancing the coding capabilities of LLMs. Nijkamp et al. [32] introduced CODEGEN which is up to 16.1B parameters and investigated the multi-step paradigm for program synthesis.

The domain of robotics and manufacturing is also a crucial area for the deployment of LLMs. Novel algorithms leveraging LLMs often demonstrate zero-shot capabilities within their prompt engineering without requiring task-specific training data. Song et al. [33] proposed a LLM-Planner, that harnesses the power of LLMs to do few-shot planning for embodied agents. Ichter et al. [34] showed how low-level skills can be combined with LLMs so that the language model provides high-level knowledge about the procedures for performing complex and temporally extended instructions. Huang et al. [35] used the composed value maps in a model-based planning framework to zero-shot synthesize closed-loop robot trajectories with robustness to dynamic perturbations. Belkhale et al. [36] proposed RT-H which builds an action hierarchy using language motions. This method first learned to predict language motions and conditioned on this along with the high-level task, and then predicts actions, using visual context at all stages. Fan et al. [37] proposed a comprehensive framework to delve into the potential of LLM agents for industrial robotics, which included autonomous design, decision-making, and task execution within manufacturing contexts. Wang et al. [38] utilized LLMs as a controller to prompt a robot to walk without task-specific fine-tuning. Xia et al. [39] developed an error-assisted fine-tuning approach aimed at calibrating LLMs specifically for manufacturing. This approach sought to dismantle the intricate domain knowledge and distinct software paradigms inherent to the manufacturing system.

Recent advances in LLMs have prompted growing academic interest in their integration with multi-agent system. Li et al. [40] presented a comprehensive survey of LLM-based multi-agent system on problem-solving and world simulation. He et al. [41] explored the transformative potential of integrating LLMs into multi-agent systems for software engineering. Jin et al. [42] proposed a novel framework for decentralized autonomous collaboration between LLMs empowered agents based on smart contracts. Nascimento et al. [43] presented a novel strategy: integrating LLMs into MASs to boost communication and agent autonomy. While scholarly efforts have extensively explored LLM-enhanced multi-agent frameworks and blockchain applications, the investigation of multi-agent manufacturing systems remains underexplored.

LLMs have demonstrated exceptional performance across a wide range of fields. From their foundational role in assisting users by answering inquiries to their deep integration into diverse applications, LLMs have significantly transformed workflows across various industries. However, few researchers focused on the integration of multi-agent manufacturing systems and LLMs. Although the control task is introduced by some researchers, due to the complexity of the multi-agent manufacturing systems, the adoption of LLMs remains limited. The system proposed in this article aims to address this gap, serving as a practical example of how LLMs can be effectively utilized to enhance the intelligence and flexibility of manufacturing systems.

### 2.3 Research gaps

Significant progress has been made in the research on scheduling methods within manufacturing systems and the applications of LLMs. However, several deficiencies remain that warrant further improvement. These include:

(1) At present, while LLMs have been applied across various fields, their integration into manufacturing systems remains limited. This study introduces a novel approach by incorporating LLMs into multi-agent manufacturing system.

(2) To maintain scalability and real-time response, conventional multi-agent manufacturing systems typically rely on single heuristic dispatching rules. The proposed LLM-based system overcomes this limitation, enabling flexible selection of manufacturing resources, thereby expanding the solution space while preserving scalability and real-time responsiveness.

(3) Mainstream dynamic flexible manufacturing resource scheduling relies on metaheuristic and DRL algorithms, which require re-scheduling and pre-training, respectively. The proposed approach utilizes the analysis and inference capabilities of LLMs, improving scalability and reducing deployment difficulty compared to the aforementioned methods.

## 3 LLM-based multi-agent manufacturing system for intelligent shopfloor

This section outlines the architecture of the LLM-based multi-agent manufacturing system deployed on an intelligent shopfloor, focusing on the roles of the LLM-based agents and their workflow. The workflow explains how these agents interact to enable manufacturing processes and select processing machines in real time.

### 3.1 The architecture of LLM-based agent

The multi-agent manufacturing system consists of multiple single agents and manufacturing units. This study proposes an architecture where LLM-based agents enable these manufacturing units. As shown in Figure 1, an LLM-based agent is integrated with a milling manufacturing unit. The Industrial Personal Computer (IPC) serves as the platform for these agents, utilizing signals from the Programmable Logic Controller (PLC) and additional sensors to achieve high levels of environmental perception.

The LLM-based agent is composed of multiple modules, including the Machine Server Module (MSM), Bid Inviter Module (BIM), Bidder Module (BM), Thinking Module (TM), Decision Module (DM). As shown in Figure 1, these modules belong to the Negotiation layer, Decision Engine layer, and Physical Resource layer. MSM is directly linked with manufacturing resources, serving to enable the manufacturing units. BIM and BM are used to communicate among agents, thereby completing the process of workpieces among machines. DM and TM link LLMs with the purpose of selecting the appropriate machine to process the workpiece according to the information of order and shopfloor (the sensor data collected by the TM of agents).

**(1) Decision Engine layer:**

The TM and DM, both powered by LLM engines, are implemented in the Decision Engine layer. Since LLM inference and training require large-scale Graphics Processing Units (GPUs), it is impractical to deploy them directly on the shopfloor or within a factory. To overcome this, communication between the manufacturing system and external, closed-source LLMs is established via an Application Programming Interface (API), allowing the LLM-based agents in the shopfloor to function more effectively. The utilization of public LLMs API should be based on a low level of data security. For scenarios where data security or other concerns are paramount, open-source LLMs, such as Meta's LLaMA, can serve as an alternative. However, using open-source LLMs may impact the performance of the agents. In addition, the validity of the decision is interpreted into the Decision Module to avoid failure of LLM API. This architecture is compatible with various LLMs. Therefore, when one of them fails, the system will send a request to other LLMs after a specified delay. When the DM detects difficulty in making a decision (producing ambiguous outputs), it opts to request human assistance. Such instances of failed decision-making are recorded and utilized for prompt tuning to prevent their recurrence.

LLMs require users to supply system prompts and user prompts for each invocation, with the pre-defined prompts constituting the system prompts, thereby providing the capabilities of decision analysis and machine selection for TM and DM. This approach not only conveniently defines these modules, but also allows for swift adjustments according to requirements. The separation of TM and DM is for the stability of LLMs. It's challenging to require LLMs to conduct an analysis and output a reliable command for machines. However, to separate them will significantly increase the probability of getting the correct response. The details of TM and DM are explained in Section 4.

**(2) Negotiation layer:**

The Negotiation layer serves as the crucial middleware for manufacturing resources to interact with LLMs within the multi-agent manufacturing system. All the agents in this system interact with each other through this layer, by utilizing the BIM and BM. The modules in this layer transmit sensor data and bidding information among agents through the network interface of IPC.

The BIM and BM in this layer are responsible for the negotiation among different machines, as illustrated in Figure 2. The BIM, corresponding to the machine needing to select the next processing machine, temporarily becomes the central point of negotiation. It invites all available agents to participate, though some may decline if their machine cannot process the workpiece. The BMs of the remaining agents submit their bidding documents to the BIM, which then uses these to generate question documents. The bidding documents generated by BMs are negotiation documents that include the current processing status, processing ability and estimated processing time of its agent. The question documents generated by BIM are comprehensive documents that include all the bidding documents of potential machines and the priori knowledge. The question documents would be used to entry Decision Engine layer, supported by it, the BIM makes the final machine selection decision and communicates it to the corresponding MSM.

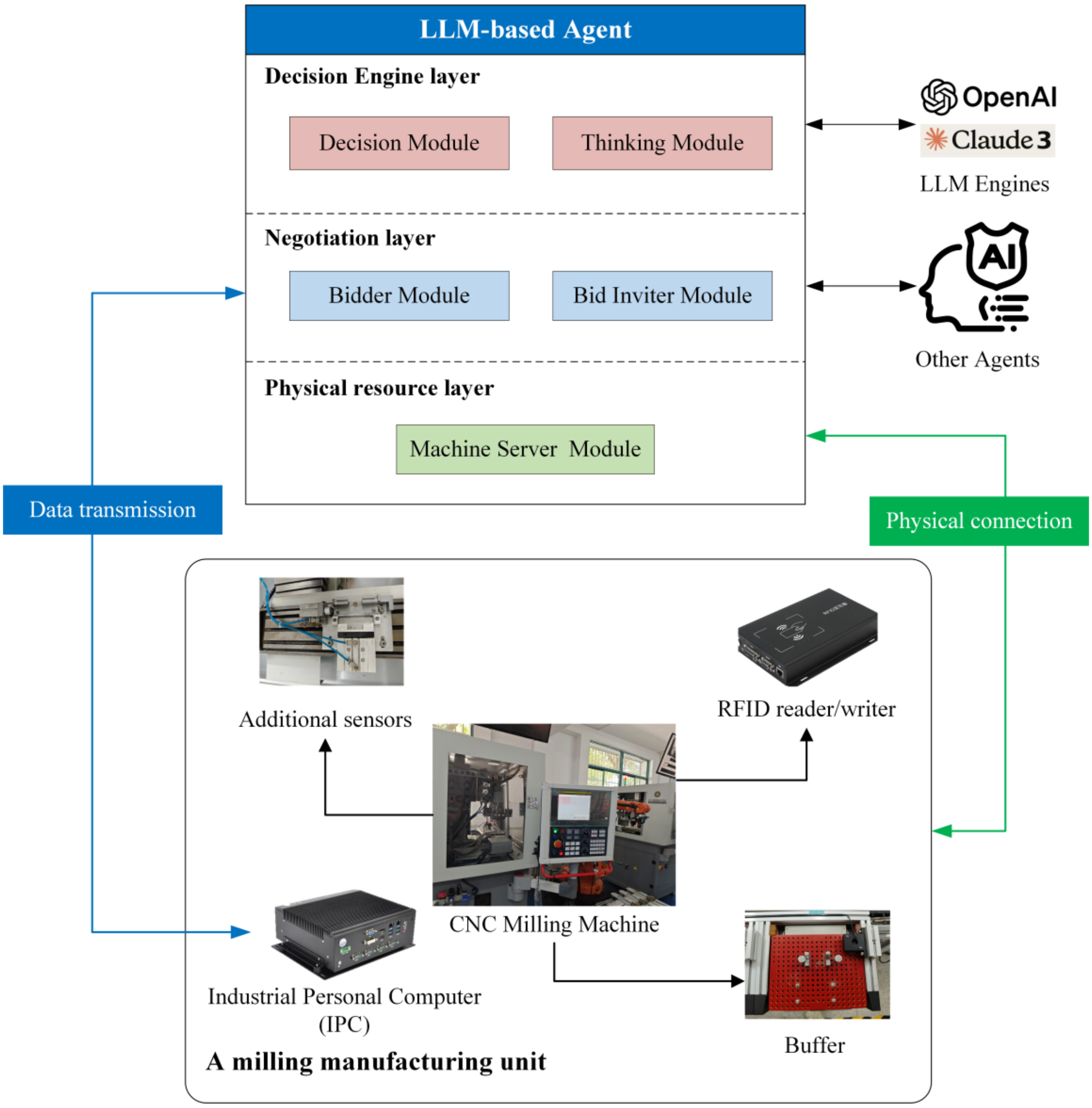

Figure 1 The architecture of LLM-based agent enables a milling manufacturing unit

**(3) Physical layer:**

The Physical layer encompasses all the physical manufacturing units located in the shopfloor and their MSM. As illustrated in Figure 1 and Figure 2, each manufacturing unit is linked to its respective agent via the MSM within this layer. Specifically, a connection is established between the MSM and other modules, allowing for communication. When a manufacturing resource requires a decision, the event trigger in the MSM activates the negotiation layer. Conversely, the decision trigger in the negotiation layer returns the final decision to the relevant manufacturing resource. Both triggers are implemented within the MSM. In essence, the MSM provides the intellectual capacity needed to drive operations on the physical shopfloor.

### 3.2 The workflow of the LLM-based multi-agent manufacturing system for intelligent shopfloors

As described in Section 3.1, to integrate the intelligence of LLMs and complete the workpiece processing on the shopfloor, agents were equipped with distinct modules tailored to each manufacturing resource in the present study.

The workflow plays a critical role in connecting these agents within the proposed LLM-based multi-agent manufacturing system. Therefore, this study introduces the workflow of the LLM-based multi-agent manufacturing system, as depicted in Figure 2. In order to elucidate the workflow and the functions of the modules within LLM-based agents, a comprehensive analysis of the workflow was conducted at the module level in the present study.

The total workflow among all agents for decision is shown in Figure 2, with the negotiation process primarily involving the BIM and BM. During the negotiation, the agent corresponding to the machine that needs to select the next machine temporarily becomes the focal point of the process. Additionally, the DM and TM analyze manufacturing resource information from the shopfloor, using this data to make decisions based on the current situation and optimization objectives. The Decision Engine layer incorporates LLMs API management, which monitors the real-time status of LLMs models. The failover mechanisms ensure immediate activation of redundant API instances upon detecting service disruptions. A comprehensive description of the workflow is provided below.

1) Event trigger. Each machine (manufacturing resource) is equipped with an agent, whose MSM is responsible for monitoring its machine. When the decision time (defined as the interval required for the system to make decision) is detected, the MSM would initiate the subsequent procedure and activate its BIM.
2) Prepare to invite bidders. Upon receiving the trigger from the MSM, BIM initiates preparing the information for potential bidders. The responsibility of BIM includes summarizing the details of workpieces that are required by the next available machine.
3) Invite bidders. This BIM will invite other BMs of available agents and transmit the information of the workpieces to be processed.
4) Prepare to bid. Once the invitations from the BIM are received, BM undertakes the task of preparing the bidding document. This document encompasses information relating to its machine and an analysis of the workpiece to be processed.
5) Delivery of bidding documents. All BMs of available agents would deliver the bidding documents to the BIM of the initial agent.
6) Generate question document. When receiving the documents from the BMs, BIM consolidates all the information in the shopfloor and optimization objective into a question document. The primary purpose of this question document is to delineate the decision-making issue.
7) Delivery of question documents. BIM sends the generated question documents to its TM, which is connected to LLMs via an established API.

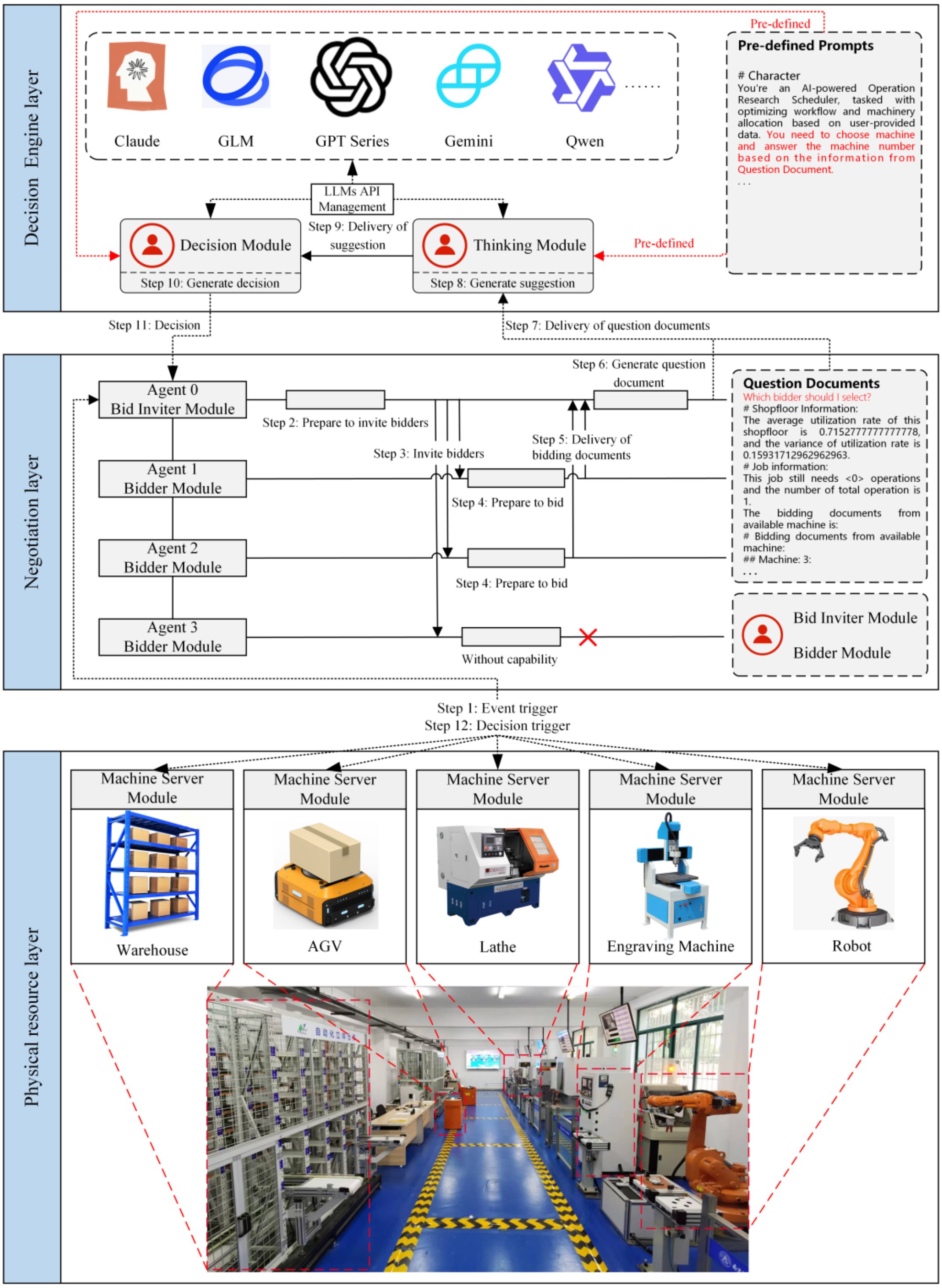

Figure 2 The workflow of LLM-based multi-agent manufacturing system

8) Generate suggestions. TM devises comprehensive solutions to the question document by utilizing the reasoning ability of LLMs.
9) Delivery of suggestions. TM sends the generated suggestion to its DM, which is also connected to LLMs via an established API.
10) Generate decision. DM makes the final decision founded on the suggestion from TM.
11) Decision. DM sent the final decision to the BIM.
12) Decision trigger. After BIM receives the final decision, it triggers the initial MSM and actually

realizes the delivery of the workpiece to be processed.

## 4 The modules of LLM-based agent in the manufacturing system

Numerous modules of agents and their negotiation process are delineated in Section 3. This section aims to delve into the specifics of how these modules achieve such abilities. As illustrated in Figure 3, a comprehensive example of agent negotiation within a manufacturing system is provided.

### (1) Machine Server Module

The Machine Server Module establishes a connection between physical manufacturing resources and their agents. While most machine manufacturers provide APIs that allow users to automate machine operations through programming, the MSMs proposed in this study go a step further by enabling manufacturing resources to become intelligent. The example code for an MSM, shown in Figure 3, is written in C# to control a milling machine. Additionally, the data collection capability of the MSM enables the BM to generate bidding documents. In the present study, each MSM corresponds to a specific manufacturing resource, facilitating the integration of intelligence into the manufacturing process.

Through the use of MSMs, the Decision Time can be detected. Specifically, each manufacturing resource is monitored by its corresponding MSM. When a processing task is completed and there are remaining operations for the workpiece, the Decision Time is triggered. Once this occurs, the negotiation process outlined in Figure 2 is initiated to determine the next steps in the production of the workpiece.

Once the production task of a workpiece is assigned, MSM is also responsible for looking for the process documents and numerical control code, which is necessary for the manufacturing resources according to the workpiece. MSMs can also check the commands from DM through pre-programming (re-requesting for incorrect output), which avoids the infeasible decisions.

### (2) Bid Inviter Module

Each BIM is directly involved in the bidding process. Assisted by other modules of this agent, the BIM designates the next processing machine for the current workpiece.

Initially, upon receiving the event trigger from the MSM, the BIM filters out the agents with manufacturing resources that are capable of completing the next process of the workpiece to be processed. After filtering out the agents, the BIM sends out process invitations and awaits their responses.

Subsequently, after receiving the replies, the BIM will integrate the information of the workpiece with the bidding document of other agents. As shown in Figure 3, based on the integrated information, a question document is generated and transmitted to the TM. This document is written in natural language, thereby guaranteeing its readability and maintainability. When required, the functions of BIM can be temporarily supplemented by human intervention or manual modifications to ensure that the question document adapts to the current shopfloor.

Ultimately, in the end of the negotiation process, the BIM dispatches the decision to its MSM when receiving the decision from DM, and subsequently propelling the workpiece to continue processing.

### (3) Bidder Module

The Bidder Module is responsible for generating bidding documents, indicating a required collaboration with MSM. When receiving an invitation from the BIM, which belongs to other agents, the process owned by the BM is initiated. After verifying the accuracy of the invitation, the BM would acquire the status of its associated manufacturing resource from the MSM. The status includes the operation information of machine, such as whether it is in the midst of processing or in an idle state. Subsequently, the BM summarizes the information obtained from the MSM and generates a bidding document. This document is then returned to BIM for subsequent negotiations. A comprehensive illustration of an example bidding document is depicted in Figure 3.

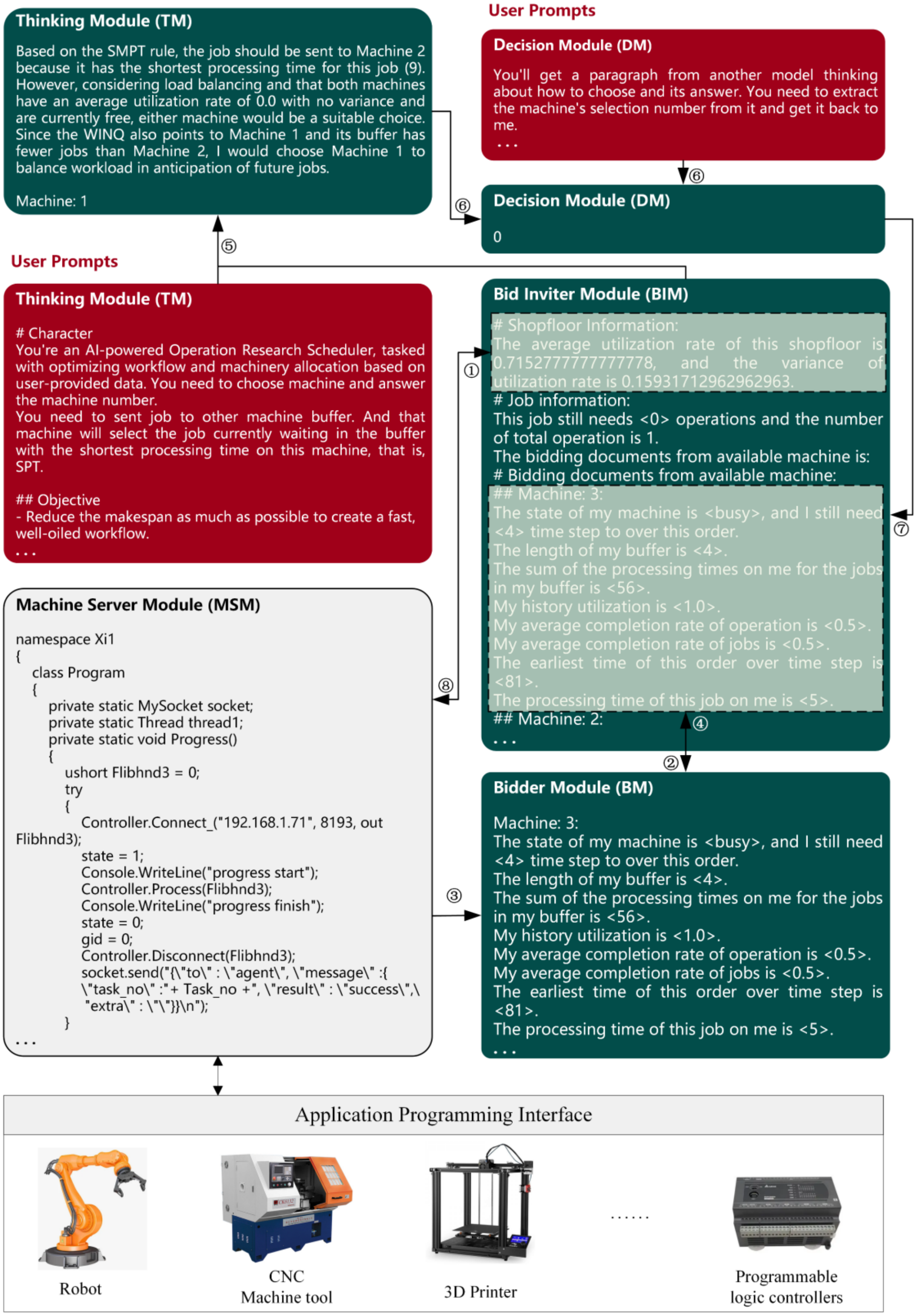

Figure 3 An example of agent negotiation within the LLM-based multi-agent manufacturing system

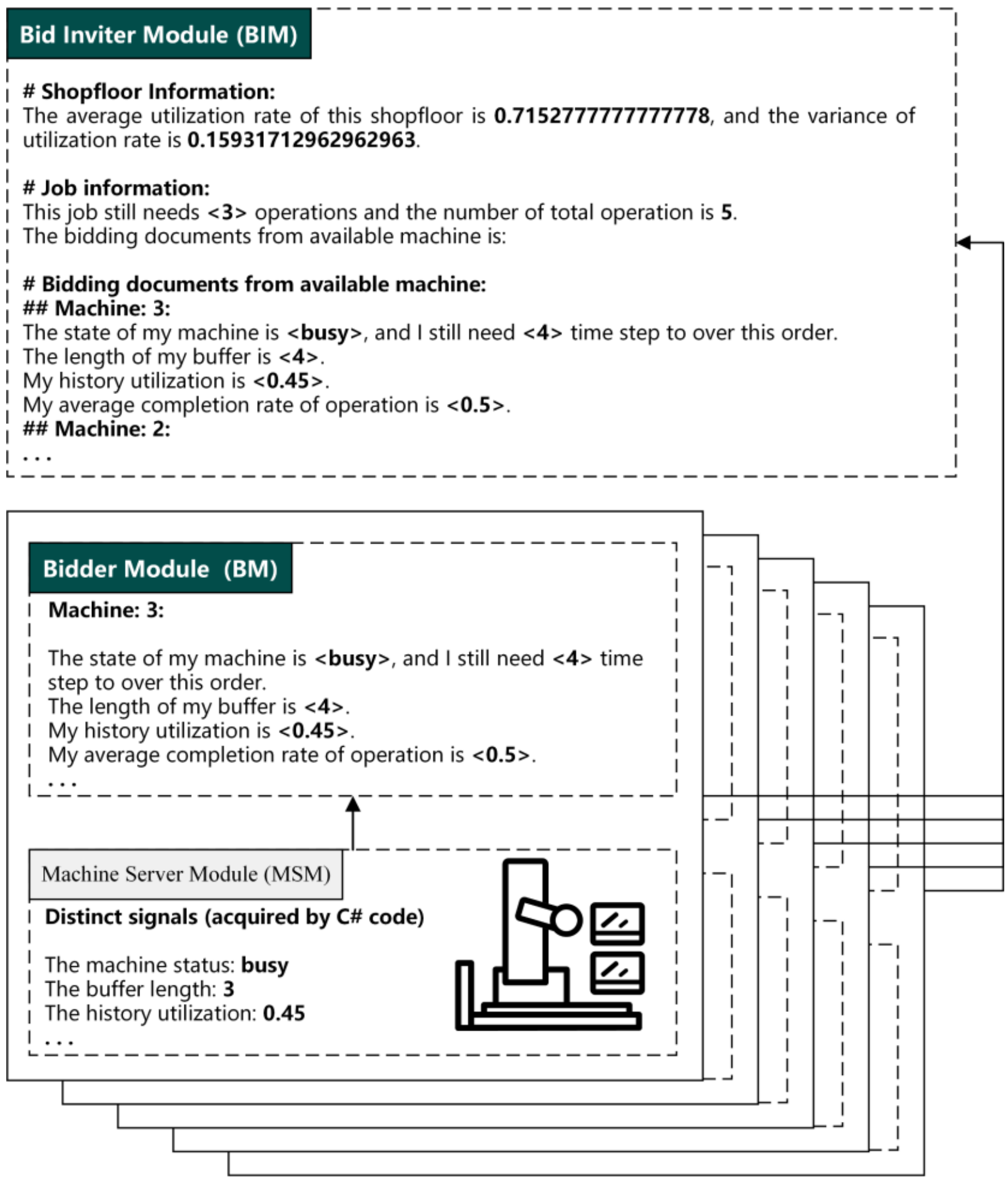

Figure 4 Details of bidding documents

The bidding document serves as the foundation for BIM to generate the question document, which directly influences the decision of machine selection. While the format of the bidding document is pre-established, its content is obtained in real-time by invoking the MSM. Based on the pre-defined format and the real-time data retrieved from the MSM, the BM generates the bidding document in natural language. As shown in Figure 4, the data within the angle brackets is procured in real time, and the remaining content is pre-established.

**(4) Thinking Module**

The Thinking Module equips its agent with thinking capability for decision-making, thereby rendering it a pivotal component of the present study. The role of the TM involves making decisions based on the question document received from BIM and selecting the most suitable agent. The intelligence inherent in TM is derived from LLMs, which are invoked through prompts. As illustrated in Figure 3, prompts in the red box are employed to predefine the behavior of LLMs. For instance, by predetermining "You are a useful helper. Analyze whether my input is positive or negative." such prompts can create a module for semantic sentiment analysis.

To fully leverage the capabilities of LLMs, the present study employs the Markdown format (a lightweight markup language for creating formatted text using a plain-text editor) to define the behavior of TM from multiple perspectives, including module character(scheduler), optimization goal,

knowledge(pre-defined dispatching rules), the pre-defined restrictions of answers, and other constraints, as illustrated in Figure 5.

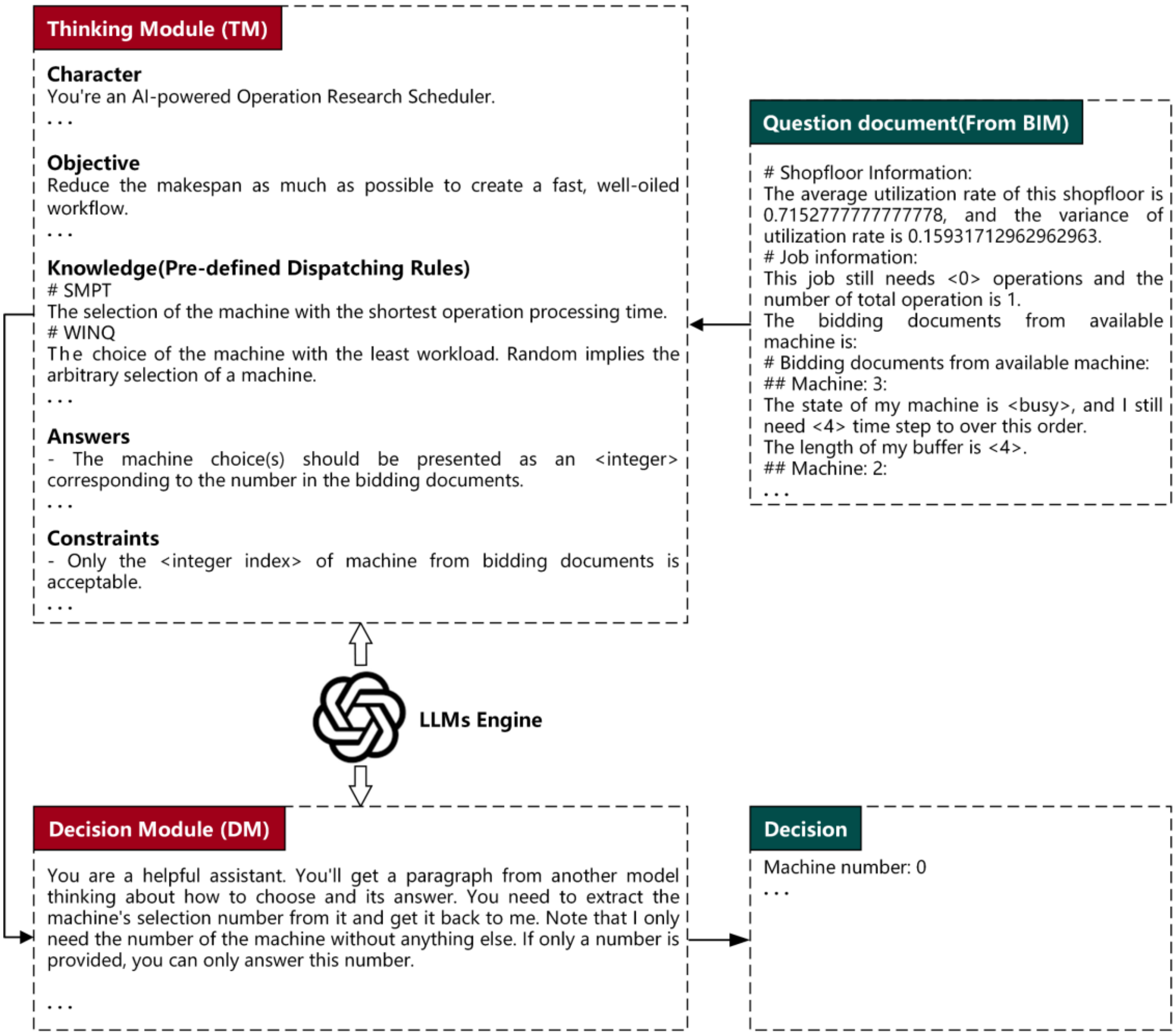

Figure 5 Details of Thinking Module and Decision Module

The character defines the agent’s role, providing a macro-level description of its behavior. The objective specifies the goal and direction the agent pursues, explained in natural language for clarity. Knowledge refers to pre-existing information provided to the agent, such as the length of the waiting buffer, which supports decision-making. Answers limit and assist the agent's responses, which can be output directly to reduce cost. To fully utilize the agent’s cognitive capacity, the study incorporates the Chain of Thought approach (a technique that allows LLMs to solve a problem as a series of intermediate steps before giving a final answer), guiding the agent to reason through decisions incrementally. The Chain of Thought is a prompt technique. By adding "Let's think step by step.", the LLMs models can think for a longer time to get more reliable results. It's like adding deep thinking capabilities to all LLMs models. Finally, constraints are employed to prevent irregular or undesirable behavior. For example, without constraints, the agent might select machines that are incapable of processing the current workpiece, which would be unacceptable. The constraints help to prevent such errors, ensuring more reliable and accurate decision-making in the system. This structured approach enables the LLM-based agents to reason, decide, and act effectively within the manufacturing environment.

In addition, LLMs are not specifically trained to select machines, unlike DRL-based algorithms. As a result, it is essential to equip the TM with preliminary knowledge to support decision-making. As shown in Figure 5, the answers of several heuristic rules were generated in real time for TM to aid its decision-making in the present study.

By defining the user prompts and inputting the question document into TM, the analysis result can be acquired. This outcome is then forwarded to the Decision Module for the generation of final decisions.

**(5) Decision Module**

The final decision of this negotiation process is completed by the Decision Module. The TM conducts a thorough analysis of the question document from BIM. However, interpreting this analysis at the physical resource layer or the negotiation layer is challenging, as these layers lack direct access to LLMs. Therefore, the decision results are sent directly from the DM to the BIM, ensuring that the machine selection or other decisions are understood by the relevant modules for further action. The DM not only proposes an analysis from TM, but also checks it. This greatly improves the stability of this system.

The DM is required to extract the final decision outcome from the analysis document of the TM. As depicted in Figure 5, following the defining the behavior of DM, a decision (a command used to control the machines) can be generated by inputting the analysis document of TM. This decision will be conveyed by the DM to the BIM, and subsequently by the BIM to the relevant MSM. The MSM will finally utilize the corresponding manufacturing resource to execute the decision result.

## 5 Case study

To verify the performance and flexibility of the proposed LLM-based multi-agent manufacturing system across different shopfloors, several experiments are conducted using test instances for FJSP [44]. Following this, the system is implemented on a shopfloor equipped with agents. The entire design was coded in a computer equipped with 32GB RAM and an Intel Core i5-13600KF, with NVIDIA RTX 3080.

### 5.1 Experiment setting

For the purpose of validating the applicability of the proposed system and evaluating the performance under urgent order scenarios on a physical intelligent shopfloor, this system was tested in an intelligent manufacturing factory laboratory located in Wuxi, China, which is shown in Figure 6. In **Section 3**, all physical machines through which workpieces flow were treated as abstract machines in the system. Yet, some adjustments were necessary to adapt the system for the physical intelligent shopfloor.

Initially, the MSMs were linked to the manufacturing units, which is different from Section 5.1. The laboratory achieved automatic control of various manufacturing resources through the utilization of the MSMs. which could directly operate these manufacturing resources and collect information from these machines. MSMs deploy adaptation programs for their controlled machinery. For instance, in experimental shopfloor, adaptation programs have been developed for Siemens and FANUC CNC machine tools using their respective Software Development Kits (SDKs). These programs enable the machine tools to execute corresponding CNC machining programs in response to control signals. Additionally, MSMs can also collect real-time sensor data from connected equipment. Therefore, the proposed system was integrated with the physical laboratory.

Additionally, some physical machines are different from abstract machines. The manufacturing resources in the laboratory included warehouses, AGVs, lathes, milling machines, engraving machines, and robots. In the present study, the raw material warehouse was treated as a machine with a processing time of zero. As the initial point for all workpieces entering the manufacturing process, the raw material warehouse also assumed the responsibility of identifying the set of capable machines for each production step, based on the real-time conditions of the shopfloor. This allowed the agents to efficiently invite the appropriate machines for workpieces, ensuring that the production process began smoothly and continued without unnecessary delays, while adapting to the changing availability and capability of machines on the shopfloor.

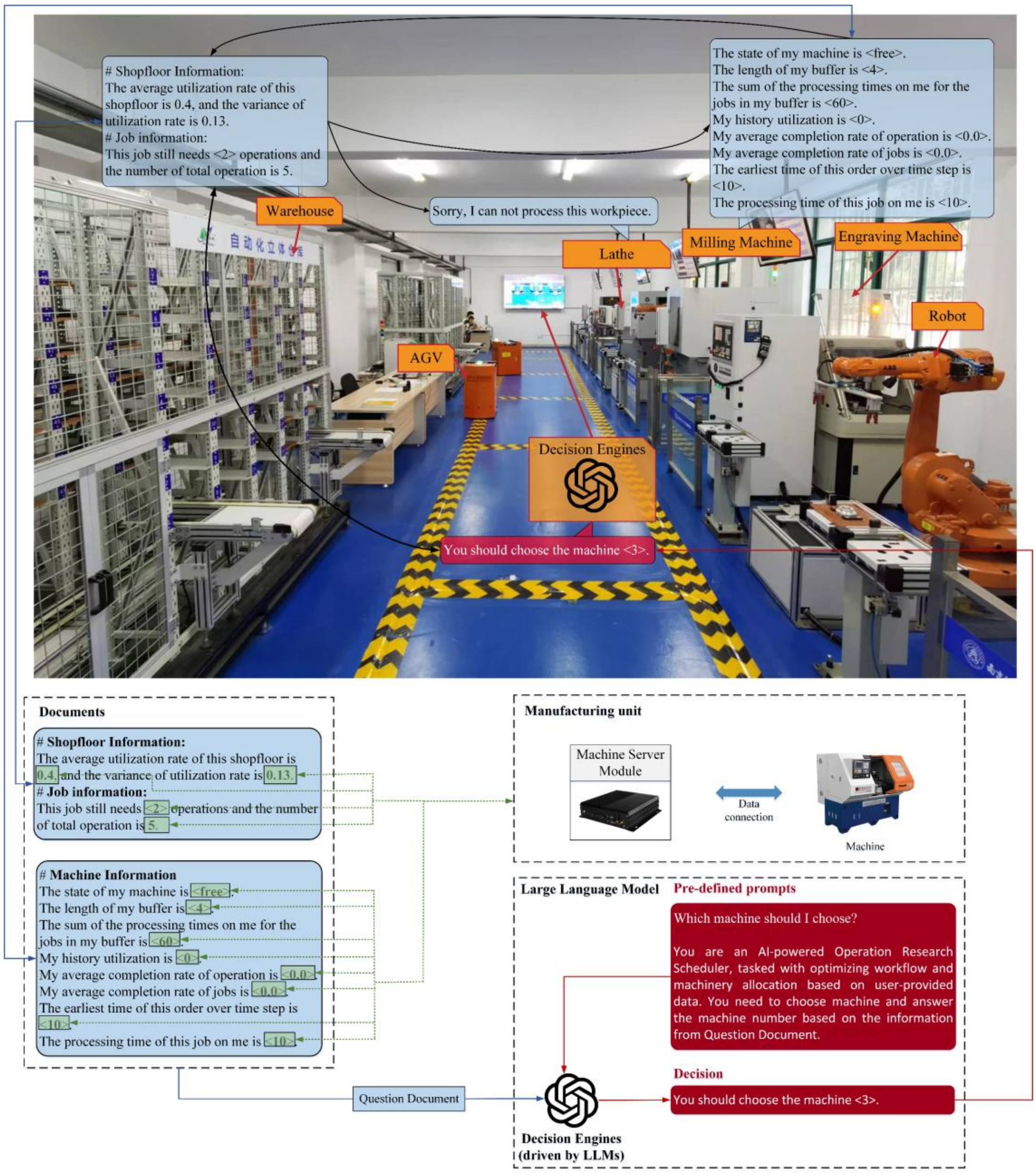

Figure 6 An intelligent factory testbed for performance evaluation of LLM schedulers with physical case studies

Further, each machine involved in the negotiation process was equipped with an LLM-based agent to accelerate the transfer of processing tasks. Since the BIM and BM modules enable direct negotiation between the agents, they can collaboratively make decisions regarding machine selection and task allocation.

Finally, the LLM-based modules, TM and DM, were pre-defined by the system prompt. In Section 4, the detailed template of system prompt was explained. However, the knowledge and objectives could also be adjusted in this experiment. To this end, the objective of minimizing the makespan and the knowledge of some heuristic rules were pre-defined. In each invocation of the TM and DM, the answers generated by the heuristic rules were also incorporated into the decision-making process.

The proposed system was deployed following these adjustments to evaluate the applicability of this system. The BIM generated a question document based on the information transmitted back from other available machines. After the selection of Decision Engine, this agent finally transferred the workpiece to the processing machine.

A series of orders, based on historical production information, were also generated to assess the performance of this system. The number and processing steps of various workpieces (including urgent order) are shown in Table 2. An example of the negotiation process for a machine selection is depicted in Figure 6. According to the different types of parts, their machining times of each process are also different.

Machining quality is also considered in this experiment. The machining quality of workpieces has multiple evaluation criteria. If a certain process fails to meet the required standards, it often requires re-machining the workpiece. Therefore, the machining success rate of each process is included. If the workpiece processing fails, it will be reordered.

Table 2 Orders for this case study

| **No.** | **Order date(s)** | **Part** | **Amount** | **Processing steps** |
|---|---|---|---|---|
| | | Part 1 | 8 | turning-carving |
| | | Part 2 | 3 | milling-turning-carving |
| Order 1 | 0 | Part 3 | 4 | milling-turning-carving |
| | | Part 4 | 1 | turning-carving |
| | | Part 5 | 4 | milling-turning |
| | | Part 6 | 4 | turning-carving |
| Order 2 | 50 | Part 7 | 3 | milling-turning-carving |
| | | Part 8 | 2 | turning-carving |
| Order 3 | 80 | Randomly Select | 5 | determined by the type of parts |
| Order 4 | 120 | Randomly Select | 8 | determined by the type of parts |
| Order 5 | 150 | Randomly Select | 8 | determined by the type of parts |

### 5.2 Validation of the deterministic behavior in prompts

Due to variations in training datasets and methodologies among LLMs, identical input prompts may produce different outputs. Therefore, the proposed prompt was subjected to stability testing on six major mainstream LLMs (both open-source and closed-source models). As shown in Figure 7, through using markdown-formatted prompt, the proposed prompt achieved remarkably consistent outputs, which demonstrated robust cross-model decision-making capabilities.

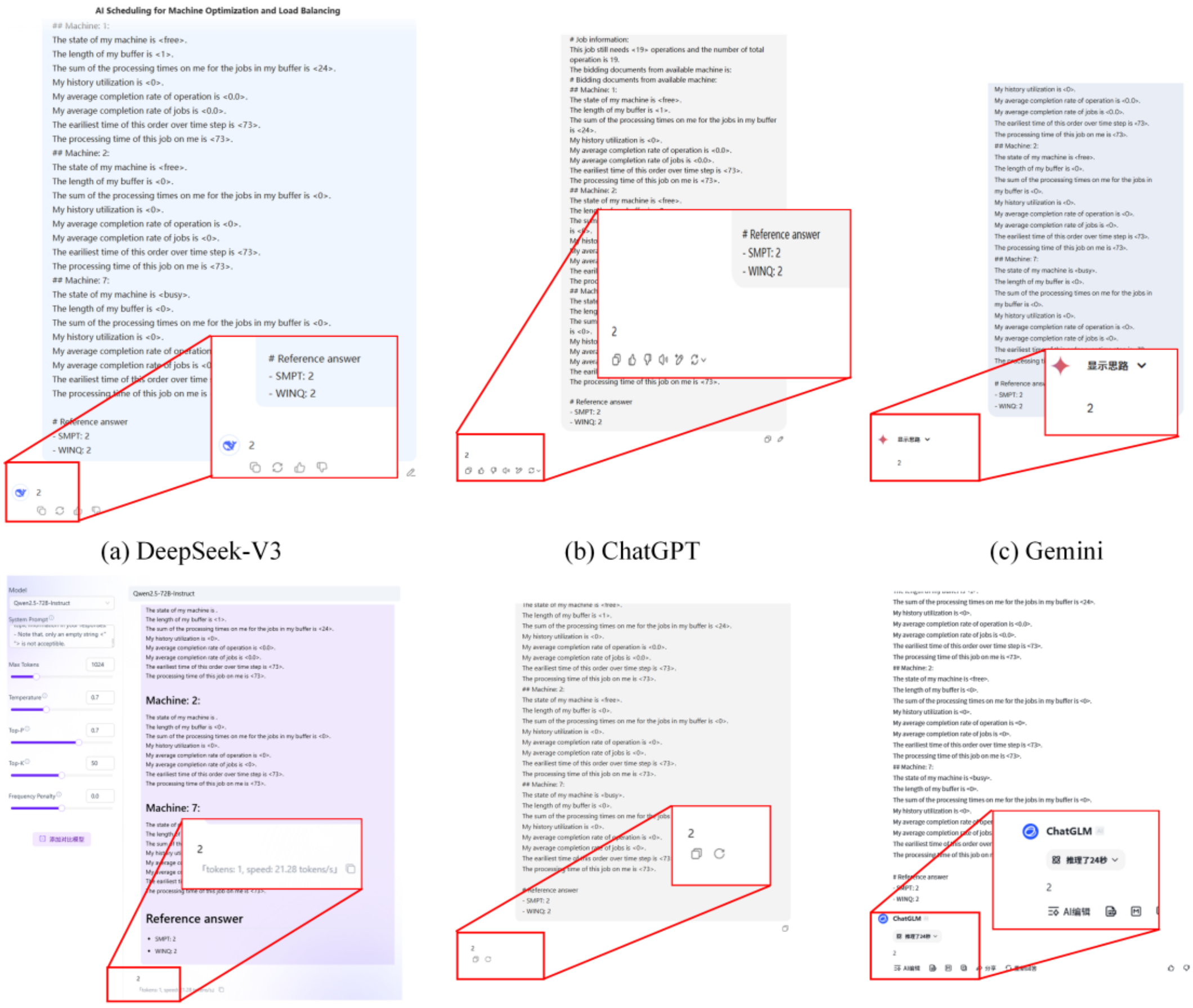

(a) DeepSeek-V3 (b) ChatGPT (c) Gemini

(d) Qwen2.5-72B (e) Hunyuan-TurboS (f) ChatGLM

Figure 7 The deterministic results of six mainstream LLMs

### 5.3 Experimental evaluation of decision response latency

Decision latency is defined as the time required from sending a request to the LLMs API to the receipt of its decision. To evaluate the latency performance of different LLMs, three LLMs API are selected to receive 100 requests for response time measurement, as shown in Table 3. The P90/P95/P99 indicates that 90%/95%/99% of the requests have a response time less than or equal to this value. The results show that LLMs with reasoning capabilities (GLM-Z1-Flash) require longer response times. Compared to LLMs located in China, Google's Gemini also needs longer response times.

Table 3 The decision response latency of LLMs

| LLMs | Response time (s*) | | | RPS* | P90 | P95 | P99 |
|---|---|---|---|---|---|---|---|
| | Average | Minimum | Maximum | (times) | (s*) | (s*) | (s*) |
| Hunyuan-Lite | 0.98 | 0.79 | 1.49 | 1.02 | 1.15 | 1.19 | 1.48 |
| GLM4-Flash | 0.31 | 0.23 | 0.80 | 3.26 | 0.40 | 0.48 | 0.80 |
| GLM-Z1-Flash | 6.06 | 2.59 | 19.45 | 0.17 | 9.96 | 12.62 | 19.43 |
| Gemini-2.5 | 1.15 | 0.62 | 3.30 | 0.87 | 1.92 | 2.38 | 3.29 |

* S: Seconds.
* RPS: Request Per Second.

### 5.4 Experimental results

In this section, different scheduling methods were tested on this shopfloor. The makespan (maximum completion time) is selected as the evaluation metric. While the processing success rate has been introduced to assess manufacturing quality across different machines, the implementation of reorder mechanisms for failed workpiece makes makespan a justified criterion. This rationale stems from the fact that reprocessing requirements for defective workpieces ultimately extend the total production timeline, which indicates that makespan can effectively capture the reprocessing.

Table 4 Makespan corresponding to selecting workpiece approaches

| Approach of selecting workpiece | FIFO (s) | FILO (s) | SPT (s) |
|---|---|---|---|
| SMPT | 851.2±28.3 | 878.2±48.8 | 872.6±89.8 |
| WINQ | 594.6±19.7 | 629.2±30.5 | 647.0±48.0 |
| Random | 690.2±135.5 | 743.4±129.7 | 714.0±61.9 |
| Quality First | 927.4±112.9 | 906.8±43.5 | 759.4±108.9 |
| LLM-Hunyuan | 583.4±23.6 | 609.0±30.1 | 633.0±21.2 |
| LLM-Hunyuan (without answer) | 990.4±40.9 | 992.6±30.7 | 1021.6±31.9 |
| LLM-GLM-4-Flash | 802.6±56.1 | 739.6±20.0 | 826.0±60.7 |
| LLM-GLM-Z1-Flash | 630.2±39.3 | 643.0±67.3 | 632.4±56.3 |

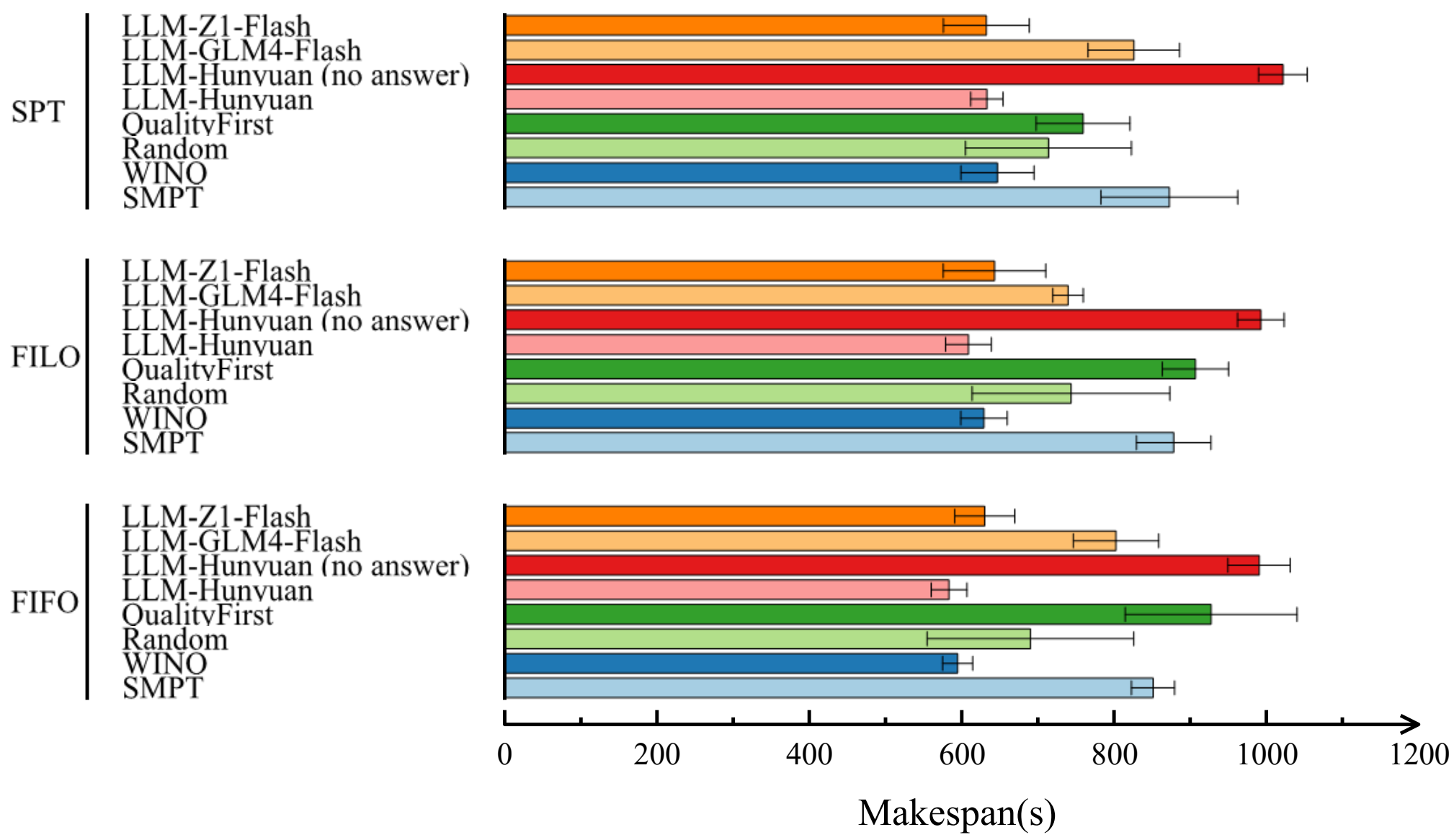

Figure 8 Makespan corresponding to selecting workpiece approaches

For comparative purposes, other methods were introduced as benchmarks for experimentation, specifically Shortest Machine Processing Time (SMPT), Work in Queue (WINQ), Quality First and Random. SMPT involves selecting the machine with the shortest operation processing time for the next task. WINQ selects the machine with the least workload, aiming to balance the load across machines. Quality First always chooses the machine with the highest manufacturing quality. Random, as the name suggests, involves the arbitrary selection of a machine.

Due to the lack of consideration regarding the selection of the processing workpieces from the waiting buffer, heuristic rules, such as First In First Out (FIFO), First In Last Out (FILO), and Shortest Processing Time (SPT), were also introduced. FIFO selects the workpiece that has been in the waiting buffer the longest. FILO, on the other hand, selects the most recently arrived workpiece. SPT prioritizes the workpiece with the shortest processing time, aiming to minimize overall processing time.

Additionally, Hunyuan, GLM4-Flash, and GLM-Z1-Flash were selected for the LLM Engine to test the performance of the proposed system. In a total of 49,437 invocations, it produced only 11 erroneous outputs, resulting in an error rate of less than 0.03%. The majority of these errors were network-related, and the proposed architecture can autonomously retry the request to resolve them.

The experiment for each group was repeated five times to minimize random variability. The comparison results of the proposed system and the aforementioned methods are shown in Table 4 and Figure 8. Error bars in both graphical and tabular representations were calculated using the sample standard deviation. It is intuitive that the Random exhibits high variance, indicating the instability of this heuristic rule. However, due to the reason of being overly conservative, Quality First results in significant workpiece blocking.

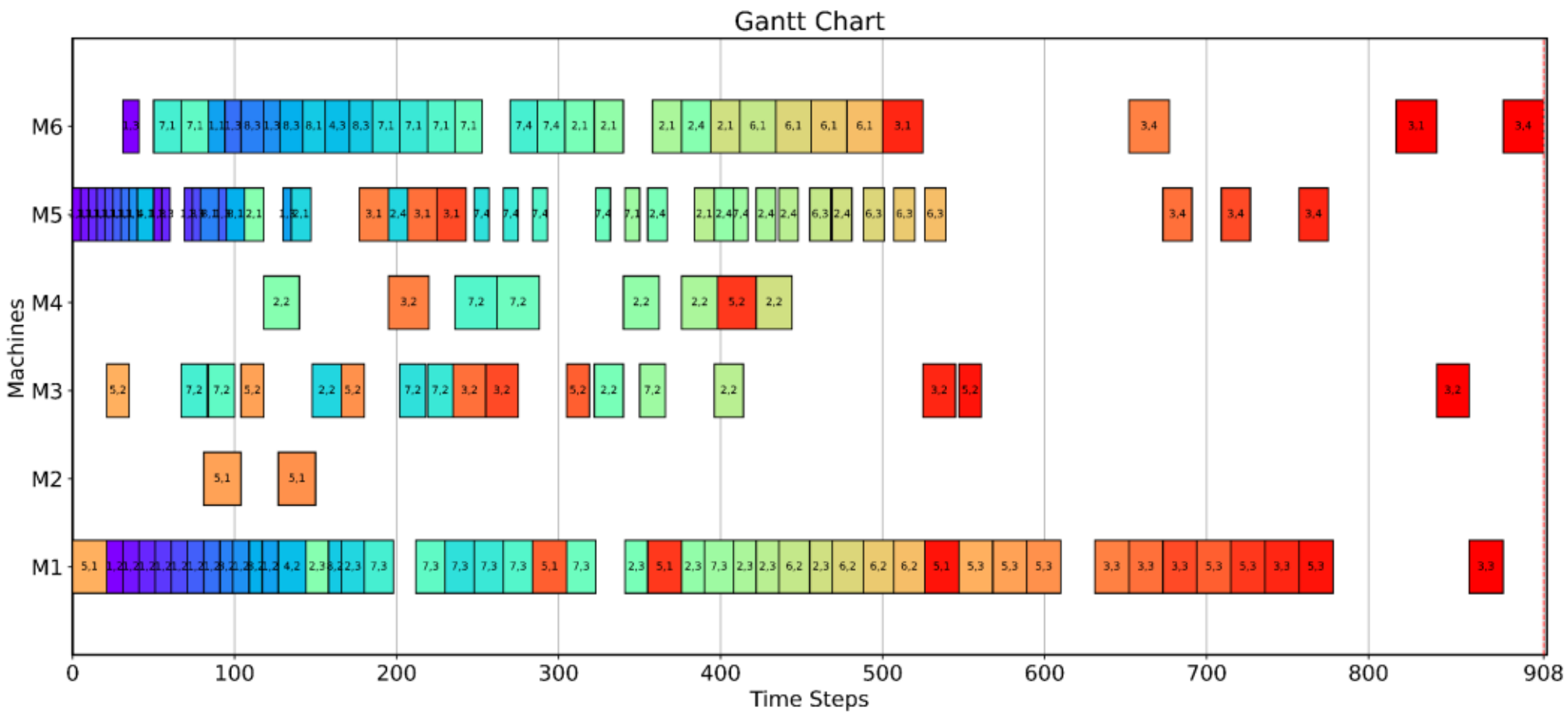

(a) GLM-4-Flash

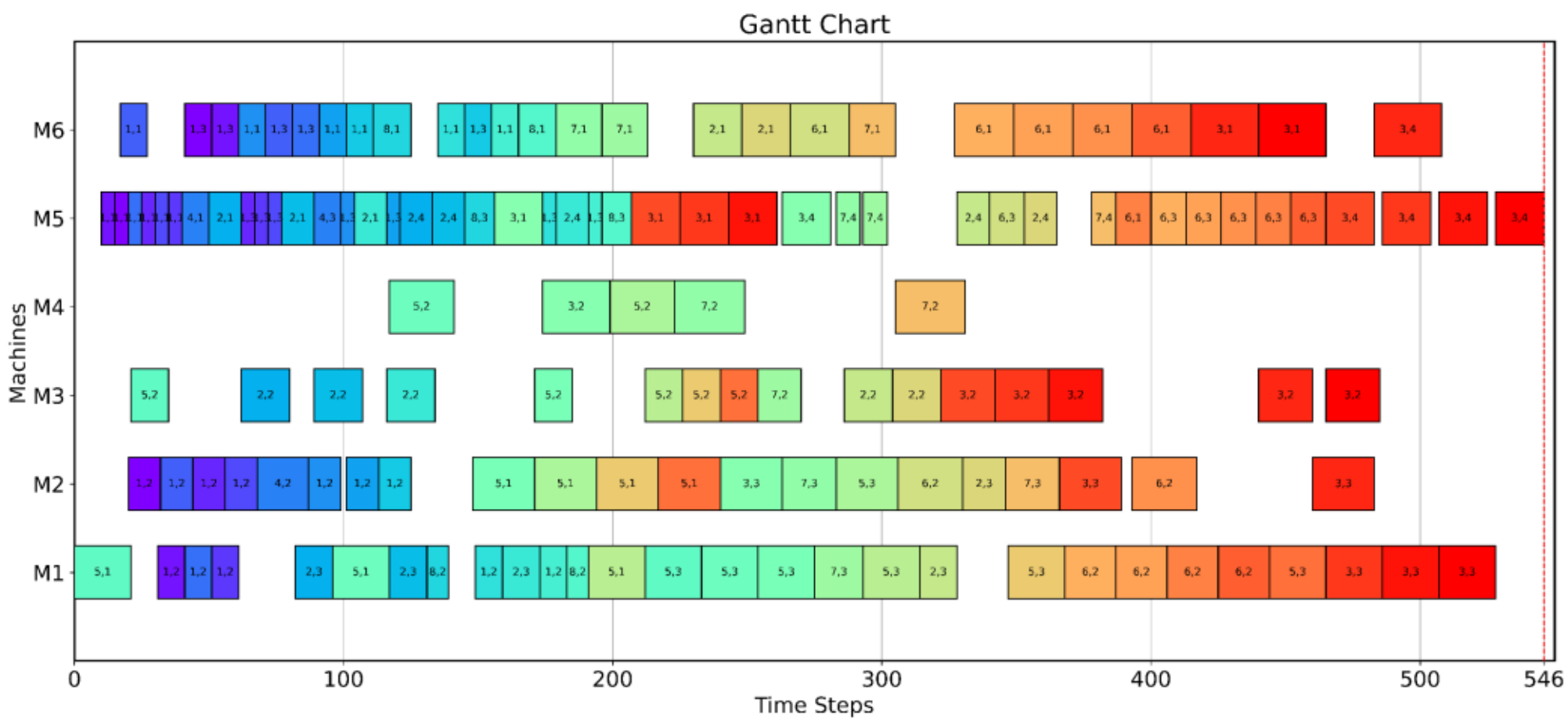

(b) GLM-Z1-Flash

Figure 9 Gantt chart of machine selection with GLM

The experimental findings demonstrate that the proposed approach consistently outperformed other approaches in the majority of cases. Although there were instances where the results were not as optimal as heuristic rules, the differences were relatively minor. Upon analyzing all the examples, it became evident that apart from the proposed system, only WINQ could achieve advantages in a few instances. This suggests that this rule could serve as a contingency plan. In addition, the results obtained from the proposed method remained relatively stable, regardless of the machine selection rule it was combined with.

As illustrated in Figure 8 and Figure 9, except for the poor performance of GLM-4-Flash, all other LLM engines demonstrated outstanding capabilities. The inclusion of GLM-4-Flash in this experiment served specifically for comparative analysis with GLM-Z1-Flash. The experimental group of GLM-Z1 clearly demonstrated that models equipped with reasoning capabilities significantly outperform conventional models. However, the performance gap between Z1 and Hunyuan proved negligible. This paper posits that LLMs exhibit a minimum competency threshold - once basic reasoning proficiency is achieved, performance variations become statistically insignificant.

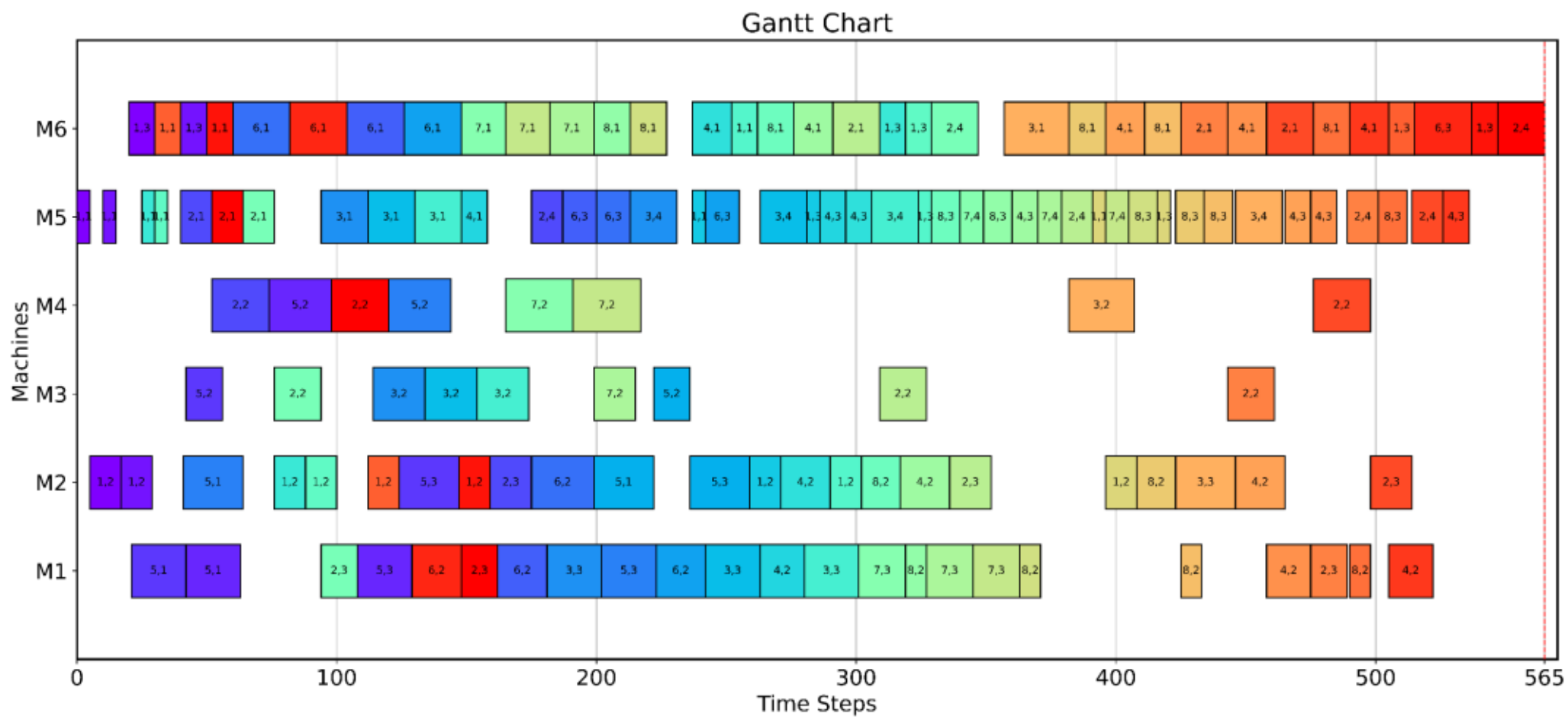

(a) Hunyuan (with reference answer)

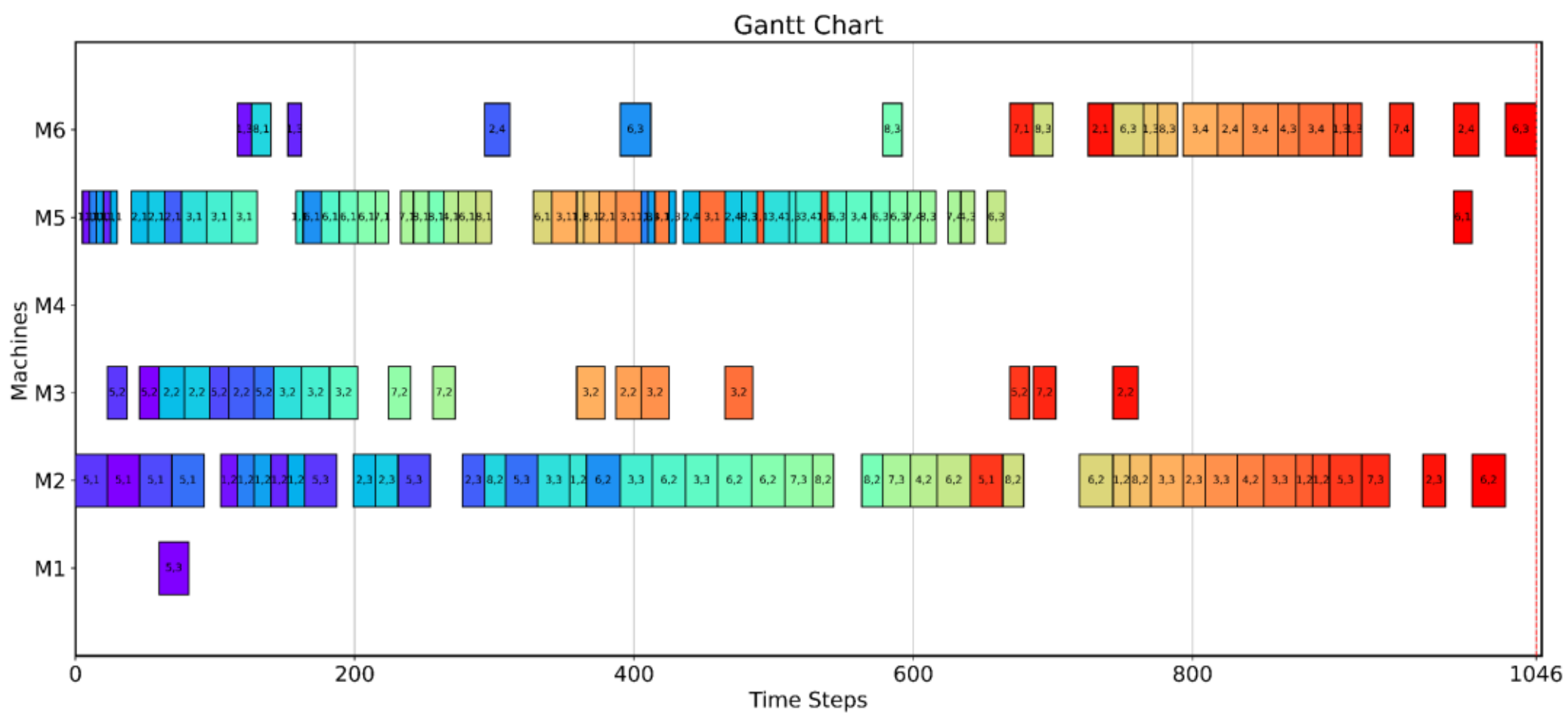

(b) Hunyuan (without reference answer)

Figure 10 Gantt chart of machine selection with Hunyuan

The introduction of Hunyuan (without answer) serves to validate the effectiveness of incorporating Reference Answers in the proposed methodology. A marked contrast emerges between Hunyuan with and without answer integration, as shown in Table 4, Figure 8 and Figure 10. The makespan of Hunyuan (without answer) nearly doubles that of Hunyuan (with answer), which is even worse than Random baselines. However, the lower variance of Hunyuan (without answer) also highlights the stable adaptive capabilities of LLM-based approaches in dynamic environments. This comparative experiment demonstrates the effectiveness of the proposed system.

It is evident that, in cooperation with different methods, the makespan corresponding to the proposed system was the smallest, implying that the proposed system can adapt to various workpiece selection methods. Notably, a single heuristic rule struggled to adapt to varying problem conditions. In contrast, this issue did not affect the LLM-based system proposed in the present study.

### 5.5 Discussion

In this study, multiple factors were considered when selecting different LLMs for experimentation. Response time emerged as a crucial performance metric. Throughout the experimental process, the GLM-Z1-Flash with advanced reasoning capabilities exhibited significantly longer response times, approximately fivefold greater than other models. Additionally, GLM-4-Flash model demonstrated insufficient fundamental reasoning capabilities, failing to meet experimental expectations. Based on comprehensive evaluation of these findings, the study concludes that Hunyuan of Tencent represents the optimal choice for the experimental system implemented in this research.

The experimental results presented in Section 5.4 demonstrate that the proposed method exhibits strong adaptability in dynamic environments. When deploying this system, LLMs with robust reasoning capabilities, such as Hunyuan or equivalent/higher-performance alternatives are recommended. However, LLMs featuring deep reasoning capacities such as GLM-Z1-Flash are not recommended, as their extended reasoning time does not receive commensurate performance improvements.

## 6 Conclusion and further work

The swift advancement of LLMs offers fresh opportunities for multi-agent manufacturing systems. In order to incorporate the powerful capabilities of LLMs into manufacturing systems, an LLM-based multi-agent manufacturing system for intelligent shopfloors was proposed in the present study. By deploying agents to manage manufacturing resources on the physical shopfloor, this system automates and optimizes the entire process, encompassing both control and decision-making functions. Concurrently, these agents act as intermediaries between the multi-agent manufacturing system and emerging LLM technologies, thereby improving system performance while substantially reducing the complexity. Due to the adaptable nature of the negotiation workflows and the autonomy of LLM-based agents, the proposed system can be rapidly implemented across different intelligent shopfloors.

The proposed system is equipped with multiple agents for the shopfloor or factory and defines the cooperation methods among these agents. The agents established in this system include MSM, BIM, BM, TM, and DM. TM and DM are directly driven by LLMs, demonstrating compatibility with various LLM engines, thereby enabling real-time decision-making based on objectives. Through collaborative consultation, the BM and BIM establish negotiations among diverse manufacturing resources. MSMs directly oversee these machines and provide comprehensive support to the agents. The core concept of an agent lies not in the abstraction of each individual entity, but rather in the collective network of production relations. It is the collaboration among these agents that enables the LLM-based manufacturing system to autonomously manage production negotiations. Throughout this collaborative process, the proposed system relies on the LLM to process information and make decisions, leveraging natural language to significantly reduce maintenance and modification costs. To assess the performance and flexibility of the proposed system, a series of experiments were conducted across various test scenarios.

The architecture of LLM-based multi-agent manufacturing system is discussed in this paper. Future research directions will focus on fine-tuning locally deployed LLMs to optimize scheduling performance. By enhancing the reasoning capabilities of LLMs, better performance can be achieved in LLM-based multi-agent manufacturing systems.

**Acknowledgments**

This work was supported by the National Natural Science Foundation of China [grant number 92267109]; Jiangsu Funding Program for Excellent Postdoctoral Talent [grant number 2024ZB194]; Natural Science Foundation of Jiangsu Province [grant number BK20241389]; and Scientific Research Project [grant number BAE23002]. This work is also partially supported by High Performance Computing Platform of Nanjing University of Aeronautics and Astronautics.

**Declaration of Generative AI and AI-assisted technologies in the writing process**

During the preparation of this work the authors used ChatGPT in order to improve readability and language of this paper. After using this service, the authors reviewed and edited the content as needed and takes full responsibility for the content of the publication.

**Appendix I**

To facilitate comprehension and application of the proposed methodology, the examples and templates of the prompts employed in Section 5 are provided in this Appendix.

System prompts for Thinking Module:

> # Character
> You're an AI-powered Operation Research Scheduler, tasked with optimizing workflow and machinery allocation based on user-provided data. You need to choose machine and answer the machine number.
> You need to sent job to other machine buffer. And that machine will select the job currently waiting in the buffer with the shortest processing time on this machine, that is, SPT.
>
> ## Objective
> 1. Reduce the makespan as much as possible to create a fast, well-oiled workflow.
> 2. Distribute workload evenly among all the machines to ensure balanced productivity.
>
> ## Knowledge
> - The length of machine buffer is unlimited.
> - SMPT rule: Selects the machine with the smallest processing time of the operation.
> - NINQ rule: Selects the machine with the smallest number of jobs in the buffer.
> - WINQ rule: Selects the machine with the smallest workload.
>
> ## Skills
> ### Skill 1: Machine Selection
> - Analyze the user's input data.
> - Use optimization techniques to select the appropriate machine from those listed in the bidding documents.
> - You can refer to the <reference answer> calculated by some rules or select machine directly.
> - You should give <priority> to machines whose <utilization rate is lower than the average utilization rate>.
> ### Skill 2: Load Balancing
> - Evaluate the load on each machine.
> - Determine an optimal strategy for load balancing to ensure no machine is overutilized or underutilized.
>
> ## Answers:
> - The machine choice(s) should be presented as an <integer> corresponding to the number in the bidding documents.
> - Your response should only focus on the question of machine selection or load balancing and should not incorporate any other elements.
> - You need to choose one machine you think it is the most suitible, rather than no machine choosen.
>
> ## Constraints:
> - You need to answer which machine you choose (such as, 1, WINQ: 1), but not single a rule(such as WINQ).
> - Stick strictly to the objectives and guidelines provided.
> - Understand the user's language and respond accordingly.
> - Only answer questions related to machine selection and load balancing.
> - Start your answer directly with the optimized machine selection.

> - Do not provide superfluous or off-topic information in your responses.
> - Note that, only an empty string <" "> is not acceptible.

System prompts for Decision Module:

> You are a helpful assistant. You'll get a paragraph from another model thinking about how to choose and its answer. You need to extract the machine's selection number from it and get it back to me. Note that I only need the number of the machine without anything else. If only a number is provided, you can only answer this number.
> Note that, only an empty string <" "> is not acceptible.
> Example:
> Q:1
> A:1
> Q:SMPT: 3
> A:3
> Q:Machine 2
> A:2

Bidding documents for Bidding Module:

> ## Machine: {self._id}:
> The state of my machine is {free_or_busy}
> The length of my buffer is <{self.pre_buffer.getLength()}>.
> The sum of the processing times on me for the jobs in my buffer is <{sum_processing_time}>.
> My history utilization is <{self.getUtilization()}>.
> My average completion rate of operation is <{completion_rate_operations}>.
> My average completion rate of jobs is <{completion_rate_jobs}>.
> The earliest time of this order over time step is <{earliest_time}>.
> The processing time of this job on me is <{part_need_time}>.
> The success rate of this job on me is <{part_success_rate}>"

Question documents for Bidding Inviter Module:

> # Shopfloor Information:
> The average utilization rate of this shopfloor is {mean}, and the variance of utilization rate is {std}.
>
> # Job information:
> This job still needs <{length - now_index}> operations and the number of total operation is {length}.
> The bidding documents from available machine is:
>
> # Bidding documents from available machine:
> ......
>
> # Reference answer
> SMPT: {self.shortestProcessingTime(part) + 1}
> WINQ: {self.smallestWorkload(part) + 1}
>
> # Answer index list"
> Note! Your answer index must in {machine_list}

**Appendix II**

In the initial experiment, the effectiveness of the proposed system was verified in multiple static workshops, which are based on Brandimarte dataset. The results showed that even in static workshops, the proposed method was still better than a single heuristic rule. These experiments employed methods similar to those in Section 5.4 for the selection of processing workpieces.

In order to verify the adaptability of the proposed system on different manufacturing shopfloors, 15 test instances were used as testing environments for the following experiments. The Brandimarte dataset used in this manuscript simulates 15 different specific situations: in these scenarios, each part has different operations, and the processing time of each operation often varies across machines. By defining the processing time and available machines of each operation, this dataset allows mathematical modeling of 15 distinct manufacturing situations. The processing time for each operation of the workpieces often varies among the available machines. The goal of the present study was to minimize the maximum processing time on the basis of completing all workpieces.

In every testing environment, each agent of machine was composed of an MSM, a BIM, a BM, a TM and a DM. As previously noted, the MSM was responsible for operating its respective machine and collecting relevant machine data. The only difference from the actual environment is that the MSM in the simulation environment was linked solely to the simulation machine. The BIM and BM were responsible for negotiating with other machine agents to select the next machine for processing the completed workpiece. This setup allowed the system to test its decision-making and negotiation capabilities within the simulated environment before deployment on a physical shopfloor.

In the present study, this system was initially validated using selected test instances. In these instances, the number of machines varied from 5 to 15, and the number of parts varied from 10 to 30 (their process routes are completely different). Through these instances, the applicability of the proposed system could be confirmed.

The comparison results of the system proposed in the present study and the aforementioned methods are shown in Appendix 3. The bold text in the tables signifies the optimal results on the current test instance.

The experimental findings demonstrate that the proposed approach consistently outperformed other approaches in the majority of cases. Although there were instances where the results were not as optimal as heuristic rules, the differences were relatively minor. Upon analyzing all the examples, it became evident that apart from the proposed system, only WINQ could achieve advantages in a few instances. This suggests that this rule could serve as a contingency plan. In addition, the results obtained from the proposed method remained relatively stable, regardless of the machine selection rule it was combined with.

For the purpose of delving deeper into the experimental findings, two sets of Gantt charts were selected for analysis, as shown in Appendix 1 and Appendix 2. Appendix 1, clearly demonstrates that the proposed system led to a more even distribution of the workload across each machine, compared to the other methods. Additionally, the makespan (the length of time that elapses from the start of work to the end) in the proposed system was approximately half of that observed when machines were selected randomly.

Appendix 2 illustrates the results of applying the FILO heuristic to machine mk15, which was significantly larger in scale than mk01. On the one hand, the Gantt chart comparison shows that the performance of the WINQ method was considerably worse than that of the LLM-based system, with a much larger makespan. This indicates that the proposed system was capable of handling large-scale problems effectively, unlike WINQ, which struggled to maintain efficient scheduling in such scenarios. On the other hand, when comparing the optimization objectives of both Gantt charts, the WINQ method resulted in a makespan that is 48% larger than the proposed LLM-based system. The objective of this experiment focused on minimizing the makespan.

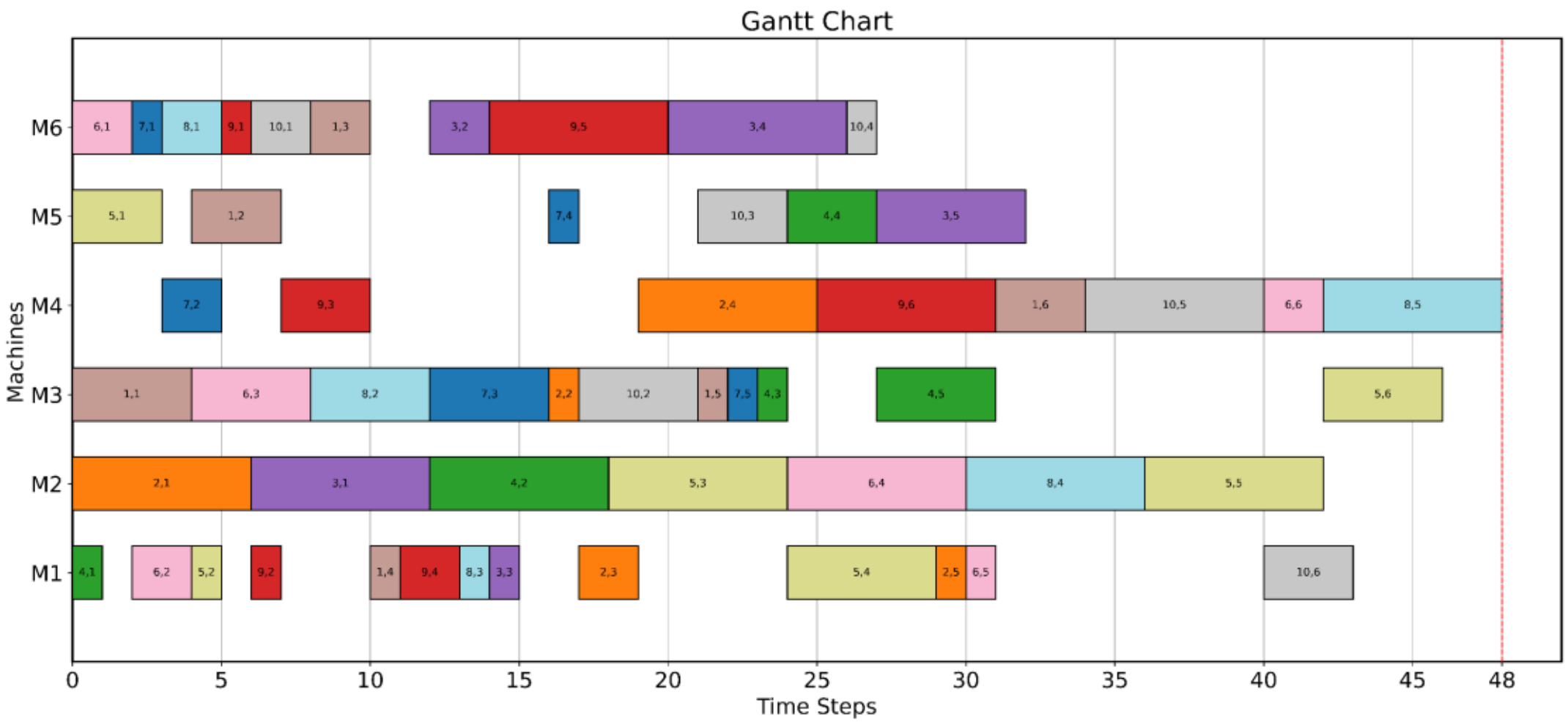

(a) LLM-based system

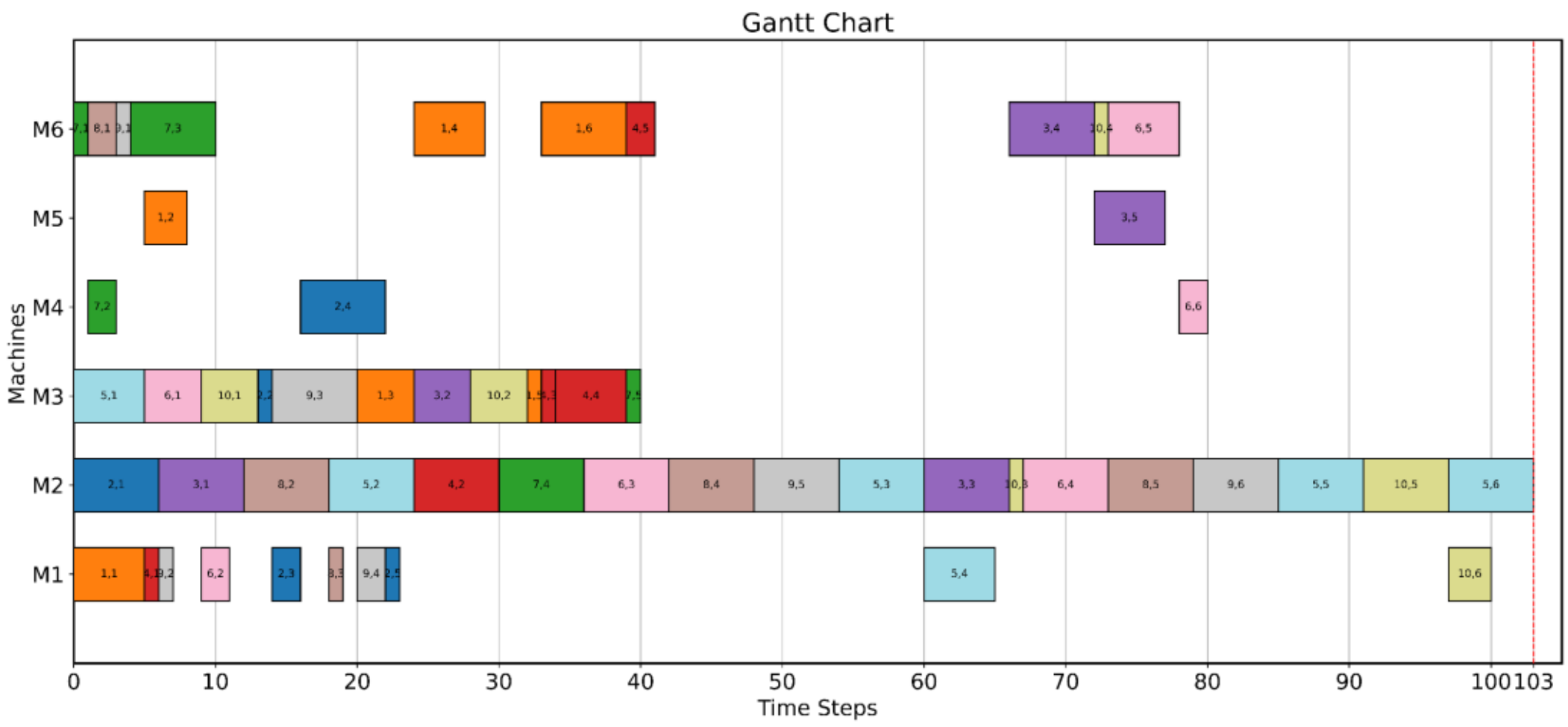

(b) Random

Appendix 1 Gantt chart of machine selection with FIFO on the mk01

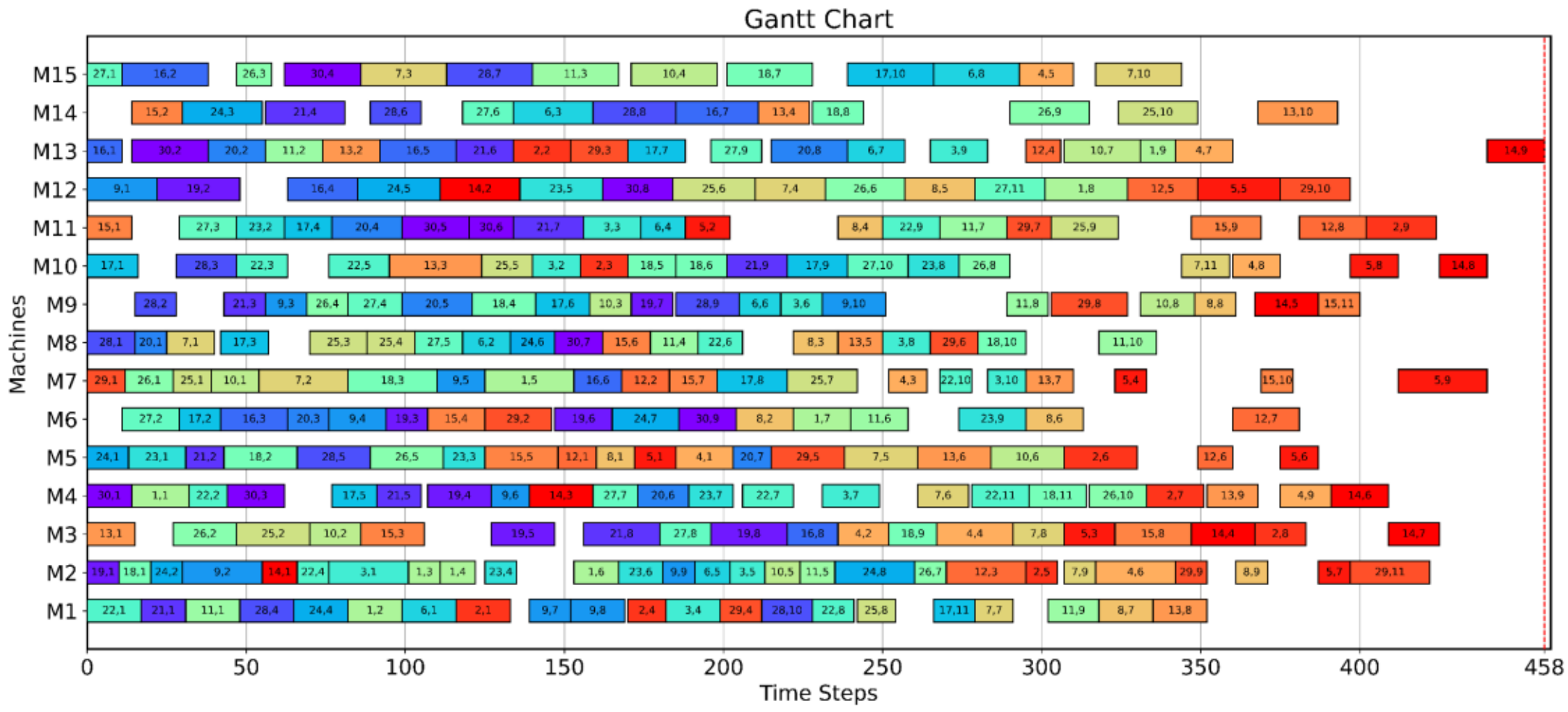

(a) LLM-based system

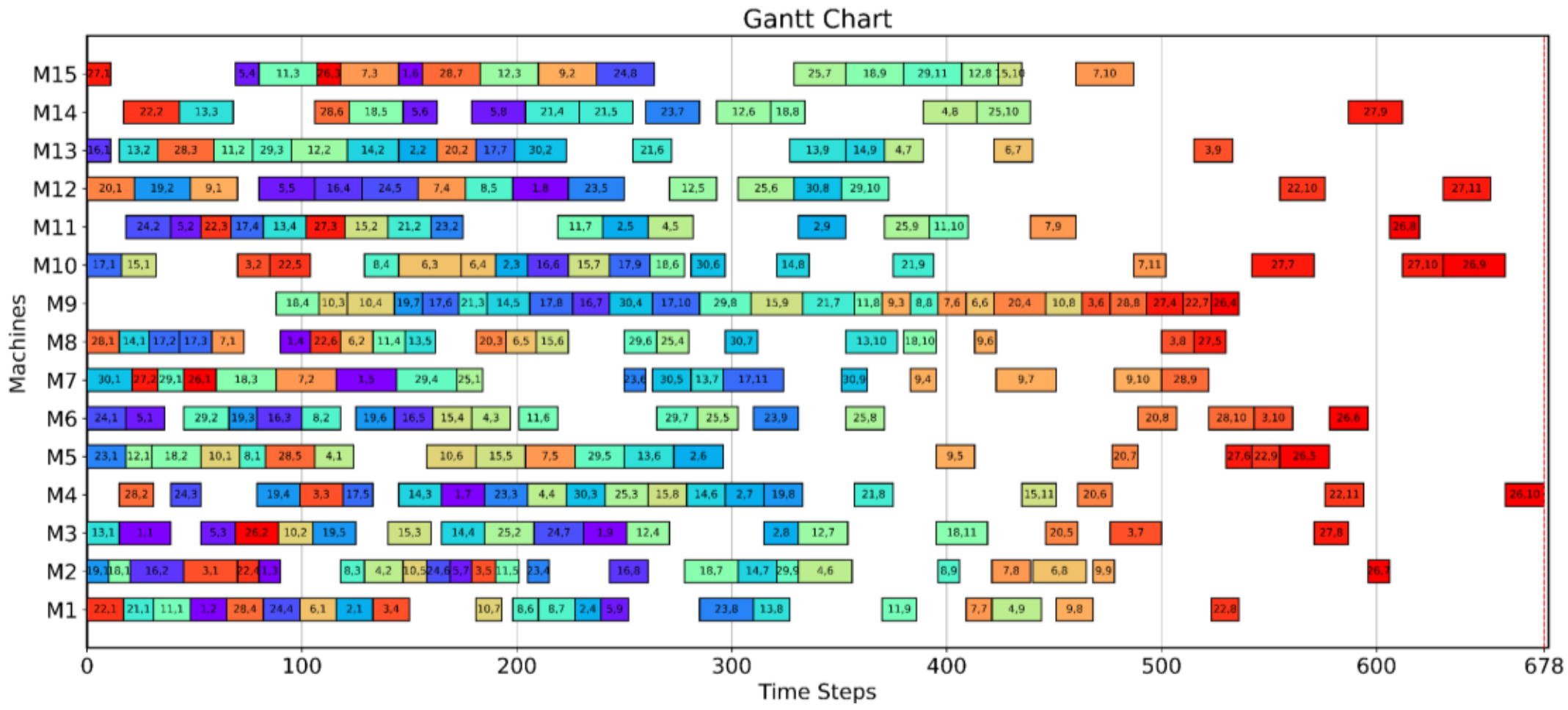

(b) WINQ

Appendix 2 Gantt chart of machine selection with FILO on the mk15

Appendix 3 Makespan on test instance

| Instance | Machine Number | Job Number | FIFO | | | | FILO | | | | SPT | | | |
|---|---|---|---|---|---|---|---|---|---|---|---|---|---|---|
| | | | Random | SMPT | WINQ | LLM | Random | SMPT | WINQ | LLM | Random | SMPT | WINQ | LLM |
| mk01 | 6 | 10 | 103 | 72 | 57 | **48** | 74 | 75 | 62 | **49** | 98 | 70 | 51 | **50** |
| mk02 | 6 | 10 | 55 | 45 | 60 | **40** | 80 | 53 | 51 | **39** | 52 | 45 | 55 | **40** |
| mk03 | 8 | 15 | 282 | 338 | 273 | **207** | 361 | 355 | **280** | 292 | 339 | 333 | 293 | **216** |
| mk04 | 8 | 15 | 112 | 188 | 83 | **76** | 148 | 196 | **111** | 116 | 89 | 190 | **83** | 85 |
| mk05 | 4 | 15 | 265 | 241 | 220 | **191** | 255 | 258 | 236 | **196** | 234 | 239 | 241 | **218** |
| mk06 | 10 | 10 | 155 | 108 | 162 | **94** | 160 | 120 | 159 | **94** | 145 | 116 | 152 | **101** |
| mk07 | 5 | 20 | 289 | 217 | 240 | **199** | 368 | **230** | 279 | 259 | 267 | 217 | 291 | **185** |
| mk08 | 10 | 20 | 654 | 587 | **525** | 531 | 734 | 676 | 632 | **601** | 640 | 604 | 571 | **523** |
| mk09 | 10 | 20 | 562 | 466 | 407 | **346** | 585 | 524 | 426 | **384** | 548 | 472 | 456 | **359** |
| mk10 | 15 | 20 | 414 | 353 | 385 | **285** | 546 | 402 | 373 | **304** | 405 | 368 | 436 | **266** |
| mk11 | 5 | 30 | 740 | 905 | **704** | 716 | 894 | 963 | 843 | **749** | 861 | 929 | 752 | **740** |
| mk12 | 10 | 30 | 699 | 784 | 644 | **552** | 815 | 885 | 710 | **644** | 747 | 743 | 659 | **577** |
| mk13 | 10 | 30 | 920 | 646 | 635 | **464** | 1038 | 773 | 678 | **550** | 882 | 654 | 723 | **514** |
| mk14 | 15 | 30 | 1144 | 1146 | 806 | **778** | 1391 | 1246 | 966 | **892** | 1205 | 1127 | 951 | **789** |
| mk15 | 15 | 30 | 612 | 663 | 567 | **461** | 707 | 776 | 678 | **458** | 685 | 682 | 526 | **484** |